# IndicNLG Benchmark: Multilingual Datasets for Diverse NLG Tasks in Indic Languages

**Aman Kumar**[1*]   **Himani Shrotriya**[2*]   **Prachi Sahu**[3*]   **Raj Dabre**[4]
**Ratish Puduppully**[5]   **Anoop Kunchukuttan**[6]   **Amogh Mishra**[7]
**Mitesh M. Khapra**[8]   **Pratyush Kumar**[9]
IIT Madras[1,2,3,6,8,9],   Columbia University[7]
National Institute of Information and Communications Technology[4]
Microsoft[6,9]   University of Edinburgh[5]   AI4Bharat[6,8,9]
[1]cs20m010@smail.iitm.ac.in   [2]cs20m024@smail.iitm.ac.in
[3]cs20m047@smail.iitm.ac.in   [4]raj.dabre@nict.go.jp
[5]r.puduppully@sms.ed.ac.uk   [6]ankunchu@microsoft.com
[7]amogh.mishra@columbia.edu   [8]miteshk@cse.iitm.ac.in
[9]pratykumar@microsoft.com

## Abstract

Natural Language Generation (NLG) for non-English languages is hampered by the scarcity of datasets in these languages. In this paper, we present the IndicNLG Benchmark, a collection of datasets for benchmarking NLG for 11 Indic languages. We focus on five diverse tasks, namely, biography generation using Wikipedia infoboxes, news headline generation, sentence summarization, paraphrase generation and, question generation. We describe the created datasets and use them to benchmark the performance of several monolingual and multilingual baselines that leverage pre-trained sequence-to-sequence models. Our results exhibit the strong performance of multilingual language-specific pre-trained models, and the utility of models trained on our dataset for other related NLG tasks. Our dataset creation methods can be easily applied to modest-resource languages as they involve simple steps such as scraping news articles and Wikipedia infoboxes, light cleaning, and pivoting through machine translation data. To the best of our knowledge, the IndicNLG Benchmark is the first NLG benchmark for Indic languages and the most diverse multilingual NLG dataset, with approximately 8M examples across 5 tasks and 11 languages. The datasets and models are publicly available [1].

## 1  Introduction

NLG is the process of generating textual output (Gatt and Krahmer, 2018). Initial work on NLG focused on tabular input (Reiter and Dale, 1997) but in a general setting the input can also belong to one or more modalities such as text, images, videos, audio, etc. NLG progress had been hindered by data scarcity for different tasks and languages but, recently, the increasing availability of large scale datasets (Narayan et al. 2018; Wiseman et al. 2017; Lebret et al. 2016), along with the advancements in neural networks pretrained on large amounts of text (Lewis et al., 2020a; Raffel et al., 2020) have led to substantial progress in NLG.

Most of the aforementioned progress is for European languages and especially for English, mainly because it is the lingua franca, making it easy to obtain data for it (Bender, 2019). However, English is not the native language for a vast majority of the world's population, which tend to use their country's or region's native languages on a daily basis. India, with its population of 1.4 billion people[2] (18% of the world population) is a quintessential example where only 10% of the population speaks English whereas a significant portion of the remaining 90% speaks one or more of the 22 'scheduled' Indian languages listed in the Constitution of India[3]. It is not surprising that most people in India tend to consume literature and media in Indian languages rather than English. Therefore, we believe that it is important to focus on Indic NLG, which lacks datasets for diverse NLG tasks.

Given that there is no existing or widely used datasets for diverse Indic NLG tasks[4], this paper aims to fill this gap, via the IndicNLG Benchmark where we create new datasets for 11 Indic languages. The 11 languages belong to two language families: Indo-Aryan and Dravidian. Dravidian languages are agglutinative, while Indo-Aryan languages are mostly not. They differ in many other aspects like gender agreement, core vocabularies, etc. The word order is the same (SOV). While the two families have their distinct core vocabularies, there exist shared vocab between these families on account of borrowings. Many other similarities are also seen due to convergence of properties over time, hence the two language families are part of the Indian subcontinent linguistic area (Emeneau, 1956).

---

\* Equal contribution.

[1]https://ai4bharat.iitm.ac.in/indicnlg-suite

[2]https://en.wikipedia.org/wiki/Demographics_of_India
[3]https://en.wikipedia.org/wiki/Languages_with_official_status_in_India
[4]Machine translation has been widely studied for Indic languages (Ramesh et al., 2022), but in this paper our interest lies in monolingual NLG tasks which have not been explored deeply for Indic languages.

| Task | Languages | Communicative Intent | Input Type | Size |
|---|---|---|---|---|
| **Biography Generation** | L-{gu, mr} | One-sentence biography | key-value pairs | 57K |
| **Headline Generation** | L | News article headlines | news article | 1.31M |
| **Sentence Summarization** | L | Synonymous compact sentence | sentence | 431K |
| **Paraphrase Generation** | L | Synonymous sentence | sentence | 5.57M |
| **Question Generation** | L | Question leading to answer given context | context-answer pairs | 1.08M |

Table 1: A summary of the 5 tasks and 11 languages (L) covered by IndicNLG Benchmark, where L={as, bn, gu, hi, kn, ml, mr, or, pa, ta, te}. The communicative intent, inputs and total corpora sizes are given.

The IndicNLG Benchmark spans five NLG tasks such as biography generation using Wikipedia infoboxes, news headline generation, sentence summarization, question generation and paraphrase generation. We also train a variety of models focusing on pre-training and multilingualism to establish strong baselines for the benchmark. Our main contributions are:

**1.** We create the IndicNLG Benchmark, a collection of NLG datasets for five diverse NLG tasks spanning 11 languages from the two major language families (Indo-Aryan and Dravidian) in India. Table 1 summarizes key characteristics of the benchmark.

**2.** It is the largest and linguistically most diverse multilingual NLG dataset, comprising a total of 8.5M examples across 11 languages and 5 tasks (~55K to ~5.57M examples for a task-language pair), opening up possibilities for multilingual NLG research.

**3.** We provide strong baselines for all tasks and languages by leveraging multilingual pre-trained models for multilingual fine-tuning, which show clear evidence of the advantage of language group specific pre-trained models compared to language agnostic ones.

**4.** We also show that the utility of our models built using mined datasets to improve performance on related NLG tasks via transfer learning.

The rest of the paper is organized as follows: Section 2 describes related work. Section 3 describes the IndicNLG Benchmark where the tasks, datasets and their creation are explained along with some important quantitative and qualitative statistics. This is followed by Section 5 where we describe the experimental settings for benchmarking the performance of our NLG models for various tasks. Section 6 contains results and analyses. We end the paper with general and task specific summaries in Section 7 and outline several future directions we plan to pursue. The appendices (Sections A, B.1, B.2, B.3, B.4 and B.5) contains additional results and analyses for interested readers.

## 2 Related Work

This paper focuses on data creation, modeling and benchmarking for NLG using pre-trained models and multilingualism.

### 2.1 NLG Benchmarks

Gehrmann et al. (2021) create a benchmark, GEM, for NLG tasks such as extreme summarization (Hasan et al., 2021; Scialom et al., 2020; Narayan et al., 2018), data-to-text generation (Gardent et al., 2017; Parikh et al., 2020; Nan et al., 2021; Dušek et al., 2020), and cross lingual summarization (Ladhak et al., 2020). In addition, they aim to establish baseline models along with automatic and human evaluations. However, the focus of GEM benchmark is predominantly English (7 out of 11 tasks). Cahyawijaya et al. (2021) propose a NLG benchmark for three Indonesian languages. Concurrent with our work, Guan et al. (2022) propose a NLG benchmark for Chinese long text NLG including two tasks. In addition, Chen et al. (2022) propose an NLG benchmark for three languages *fr*, *de* and *es* along with *en*, for three tasks of story generation, headline generation and question generation. In contrast, our IndicNLG benchmark covers 11 Indic languages and five tasks, making it the first for Indic languages as well as the most linguistically diverse NLG benchmark to the best of our knowledge. IndicNLG Benchmark complements the IndicGLUE benchmark (Kakwani et al., 2020) for Indic natural language understanding (NLU).

## 2.2 NLG Tasks

IndicNLG Benchmark focuses on data-to-text generation (Reiter and Dale, 1997; Gatt and Krahmer, 2018; Lebret et al., 2016), headline generation (Banko et al., 2000), sentence summarization (Rush et al., 2015), paraphrasing (McKeown, 1979; Meteer and Shaked, 1988; Madnani and Dorr, 2010) and question generation (Brown et al., 2005; Ali et al., 2010; Heilman, 2011; Du et al., 2017). Following previous works, we create datasets for each task automatically from structured data such as Wikipedia (Lebret et al., 2016) or news articles (Banko et al., 2000) or by leveraging parallel corpora (Madnani and Dorr, 2010) or by machine translation of an English dataset. Furthermore, we clean the datasets using regular expression and statistics based cleaning.

## 2.3 Pre-trained models and multilingualism

The availability of pre-trained models, typically trained using unsupervised approaches and monolingual data, help reduce the requirement of large amounts of (supervised) fine-tuning data for a given downstream task. In this context, T5 (Raffel et al., 2020), mT5 (Xue et al., 2021), BART (Lewis et al., 2020a), mBART-25 (Liu et al., 2020) and mBART-50 (Tang et al., 2021) are commonly used for fine-tuning. More recently Dabre et al. (2022) introduce a pre-trained sequence-to-sequence model for Indic languages, which we use in this paper. Previous research in machine translation and summarization have shown that multilingual models fine-tuned on pre-trained models tend to yield the best results (Hasan et al., 2021; Ramesh et al., 2022), a direction we also follow.

## 3 IndicNLG Benchmark

The IndicNLG Benchmark is a collection of datasets which we use to benchmark NLG performance for 5 NLG tasks spanning 11 Indic languages. In this section, we describe the datasets and their sizes.

## 3.1 Tasks and Languages Choice Criteria

Task choice depends on language coverage, task coverage and practical applications. Regarding language choice, our priority is to include as many languages as possible, where we are currently limited to 11.

| L | BG | HG | SS | PG | QG |
|---|---|---|---|---|---|
| as | 2,072 | 59,031 | 21,496 | 8,840 | 98,027 |
| bn | 7,703 | 142,731 | 21,774 | 910,445 | 98,027 |
| gu | 0 | 262,457 | 71,968 | 399,202 | 98,027 |
| hi | 9,456 | 297,284 | 112,340 | 949,507 | 98,027 |
| kn | 1,960 | 155,057 | 71,729 | 542,148 | 98,027 |
| ml | 9,351 | 20,966 | 5,955 | 781,933 | 98,027 |
| mr | 0 | 142,590 | 34,026 | 426,003 | 98,027 |
| or | 2,760 | 72,846 | 15,044 | 125,970 | 98,027 |
| pa | 6,354 | 60,635 | 39,601 | 286,704 | 98,027 |
| ta | 13,502 | 75,954 | 28,920 | 517,798 | 98,027 |
| te | 4,268 | 26,717 | 8,859 | 616,283 | 98,027 |
| total | 57,426 | 1,316,268 | 431,712 | 5,564,833 | 1,078,297 |

Table 2: Dataset sizes in number of examples for 5 tasks of WikiBio biography generation (BG), Headline Generation (HG), Sentence Summarization (SS), Paraphrase Generation (PG) and Question Generation (QG) spanning 11 languages in the IndicNLG Benchmark.

## 3.2 IndicNLG Tasks

We focus on biography generation (BG) using Wikipedia infoboxes (WikiBio), news headline generation (HG), sentence summarization (SS), paraphrase generation (PG) and question generation (QG). Dataset sizes in number of examples for each task and language are given in Table 2. Except for WikiBio, datasets are available for all 11 Indian languages of interest namely, Assamese (as), Bengali (bn), Gujarati (gu), Hindi (hi), Kannada (kn), Malayalam (ml), Marathi (mr), Odia (or), Punjabi (pa), Tamil (ta) and Telugu (te). All sizes reported are after deduplication. Due to lack of space, we only give the important details, and we encourage readers to check Appendix B.1, B.2, B.3, B.4, and B.5 for BG, HG, SS, PG and QG tasks, respectively, for further details regarding the dataset construction and cleaning process, quantitative and qualitative statistics, and examples.

### 3.2.1 Biography Generation (WikiBio)

The WikiBio task was first proposed for English, where, given the Wikipedia *infobox* of a person, the objective is to generate the first sentence of its Wikipedia page (Lebret et al., 2016). An infobox is a table containing facts in a key-value format, and the task objective is the summary of the infobox. In order to create the datasets, we crawl the Wikipedia pages of the aforementioned languages, except Marathi and Gujarati[5], preprocess and filter them to ensure high quality. The "BG" column in Table 2 gives the statistics of the final corpora. We extracted a total of 57,426 examples, with Assamese and Tamil having the least and most number of examples, respectively. The English Wikibio dataset contains 728,321 examples, which shows that our dataset, which is ~6% the size, is very low-resource.

---
[5] The number of Wikipedia pages in were too low to ensure a substantial number of validation and testing examples.

### 3.2.2 Headline Generation

Headline generation is the task in which, given an article, the objective is to generate an appropriate sentence, a title, that accurately depicts the article (Banko et al., 2000). The headline should be able to draw the reader's attention while compressing information from several hundreds of words into a single sentence. The raw data for Hindi is crawled from HTML web pages of various domains like Dainik Bhaskar, Naidunia, NDTV, Business Standard and IndiaTV to ensure content diversity. We extract document and headline pairs and filter noisy examples. For other languages, we used the Headline Prediction dataset from the 'IndicGLUE' benchmark (Kakwani et al., 2020) [6] where we chose the document as the input and the correct headline as the output.

The column "HG" in Table 2 gives the statistics of the final corpora. There are a total of 1.31M examples, with Hindi containing the most (297K) and Malayalam containing the least (20K) number of examples. Comparing this with the corresponding Indic section of the XL-sum dataset (Hasan et al., 2021) which can also be used for headline generation, we have more examples for each language. For example, we have 297K examples for Hindi, whereas XL-Sum only has 88K examples. The Indic section of XL-Sum has 167K examples and English has 329K examples, a lot smaller than our 1.31M examples.

### 3.2.3 Sentence Summarization

Sentence summarization involves compressing the information of a reasonably long sentence into a shorter, compact sentence (Rush et al., 2015; Chopra et al., 2016). Following Rush et al. (2015), we create a sentence summarization dataset where the input is the first sentence of a news article and the output is its headline. The intuition is that the first sentence in a news article often expands upon the information in the headline, which makes the headline a summary of the first sentence. We simply re-process the headline generation dataset by extracting the first sentence and headline pairs to create our sentence summarization dataset. However, not all first sentences were valid document summaries, and we discard such examples to ensure high quality.

The column "SS" in Table 2 gives the corpora statistics, where we have a total of 431K examples. The count of examples in the training set ranges from 5.9K for Malayalam (least) to 112K for Hindi (most). Although this dataset is derived from the headline generation dataset, the number of examples is far fewer than the latter, which has 1.31M examples. Nevertheless, there are more examples compared to the XL-Sum counterpart, which contains 167K examples. The Gigaword[7] corpus for sentence summarization for English (Rush et al., 2015), containing over 4M examples, however, is almost an order of magnitude larger.

### 3.2.4 Paraphrase Generation

Paraphrase generation or paraphrasing (McKeown, 1983; Barzilay and Lee, 2003) is the task of transforming a sentence into a different sentence in the same language while preserving meaning and semantics. A paraphrasing system is important as it enables generation of alternatives for a given sentence. Following Zhao et al. (2008), we use the pivoting approach to extract paraphrases from a parallel corpus. The intuition is that sentences are paraphrases if they have the same translation. To this end, we use the Samanantar corpus (Ramesh et al., 2022) which contains parallel corpora between English and all 11 Indic languages of interest[8]. Using English as the pivot language, we extract paraphrases for each language. Since this approach can lead to multiple paraphrases with the same meaning, we choose one as the input and then retain up to 5 paraphrases ordered from lexically dissimilar to similar. We hope this will enable a paraphrasing system to be learned that can generate diverse paraphrases depending on the user's needs.

In Table 2, the "PP" column gives the corpora statistics, indicating a total of 5.57M examples, making this task the most resource rich in this benchmark. Each example is a tuple. Sizes vary strongly between languages, where Hindi has 950K paraphrases and Assamese has 8,840. Comparing our dataset with the OpusParcus corpus

---

(Creutz, 2018) for 6 European languages, we are almost 3 orders of magnitude larger and have up to 5 references per example, where the latter has only 1. However, OpusParcus has been fully manually checked, whereas ours is not.

### 3.2.5 Question Generation

Question generation is the task of generating a question given some context and the answer (Du et al., 2017; Zhou et al., 2017). Question generation can be extremely useful to teachers in designing examination questions given some fixed answer. Unlike the previous tasks, creating data for this task is quite expensive, and thus we rely on machine translation. Following earlier work (Du et al., 2017; Zhou et al., 2017; Dong et al., 2019 inter alia), we start with the SQuAD training and development (Rajpurkar et al., 2016) question answering sets and repurpose them to serve as a question generation dataset. Specifically, we extract the question, its answer and the sentence containing the answer. We designate the input to question generation as the sentence along with the answer. The question serves as the output. We then translate the data into Indic languages using the IndicTrans (Ramesh et al., 2022)[9] English to Indic model. We end up with 98K examples per language and 1M total across all languages. Previous work has used IndicTrans to create testsets and found the translation to be of good quality (IndicXNLI; Aggarwal et al. 2022).

### 3.3 Summary of Datasets

In summary, the IndicNLG Benchmark contains diverse NLG tasks for 11 Indic languages that vary in their linguistic characteristic and resource availability. While the corpora are smaller than their English counterparts in some cases, they are of reasonable size for building a benchmark dataset. Moreover, the relatedness of Indic languages opens up the possibility of training multilingual generation models. In the appendices, we also report extensive metrics to quantify characteristics of the datasets. The metrics show that the datasets are as challenging (if not more) as standard English datasets for these tasks, as measured by n-gram novelty and simple baseline approaches (Please see Tables 10, 15, 20, 24 and 30 for individual dataset qualitative metrics).

---

| BG | Property A | Property B | Property C |
|---|---|---|---|
| kn | 98.4 | 6.3 | 63.6 |
| bn | 99.4 | 8.3 | 18.6 |
| hi | 98.8 | 18.9 | 45.2 |
| ml | 96.0 | 14.8 | 47.1 |

| HG | Property A | Property B | Property C |
|---|---|---|---|
| kn | 91.5 | 3.8 | 4.4 |
| bn | 91.4 | 3.9 | 9.7 |
| hi | 91.0 | 5.5 | 4.9 |
| ml | 90.5 | 2.2 | 3.3 |

| PG | 5 | 4 | 3 | 2 | 1 | 0 |
|---|---|---|---|---|---|---|
| kn | 24.5 | 15.0 | 18.2 | 8.0 | 12.0 | 22.2 |
| bn | 34.2 | 28.1 | 18.4 | 2.9 | 9.4 | 7.0 |
| hi | 39.3 | 17.8 | 18.8 | 13.4 | 7.7 | 3.1 |
| ml | 23.9 | 27.6 | 14.9 | 13.7 | 10.5 | 9.5 |

Table 3: Human evaluation of quality of WikiBio (BG), Headline Generation (HG) and Paraphrasing datasets (PG). For WikiBio and Headline Generation, Property A measures the match between the output and input, Propery B measures if output contains information inconsistent with the input, and Property C measures if the output contains information that cannot be inferred from the input. For Paraphrasing, Labels 5, 4, 3, 2, 1, 0 indicate the match between output and input from exact match (Label 5) down to unrelated pairs (Label 0).

## 4 Dataset Quality

We study the quality of our automatically created datasets by conducting a human evaluation exercise. We choose two languages from each of Indo-Aryan and Dravidian language families as representative languages *viz* we choose *hi* and *bn* from Indo-Aryan and *ml* and *kn* from Dravidian language families. We conduct this study for only three tasks out of four (WikiBio, Headline Generation and Paraphrasing) due to limited annotation budget.

We annotate 250 examples in WikiBio and Paraphrasing, and 100 examples in Headline Generation (the examples are longer here, so we annotate fewer examples to save on annotation cost). For each language, we hire two native-language annotators and pay them higher wages than the minimum hourly wages. Table 3 contains the results of our human evaluation exercise, details of which are described below.

For annotation of WikiBio and Headline Generation we follow the human evaluation setup of Hasan et al. (2021). Specifically, we ask raters to annotate for three properties.

**Property A** is *Yes* if the output and input pair are aligned.

**Property B** is *Yes* if the output contains information inconsistent with the input.

**Property C** is *Yes* if the output contains extra in-

---

formation that cannot be inferred from the input.

The numbers of Property A, B and C mean the ratios of 'Yes' given by the annotators. It is desirable that values of Property A should be high, and values for Property B and C be low. We find that Property A is greater than 90% indicating high match between the input and output example pair. Property C values are greater than 10% in WikiBio, indicating the extra information present in output, an observation also made in Hasan et al. (2021). We find that the amount of extra information present in the output of Indic WikiBio dataset is similar in nature to abstractive summarization dataset such as XSum (71.7% from Table 3 in Hasan et al. 2021). In addition, as discussed in Hasan et al. (2021), large pretrained models are able to make use of external information from texts these models were trained on. So, we believe that the extra information present in the outputs will not have a big adverse effect on the quality of the datasets. Property C values are lower than 10% in Headline Generation. Property B is relatively low (around 5%) as desired for Headline Generation; however, it is higher for WikiBio dataset.

For the annotation of Paraphrasing data, we follow the setup of Cer et al. (2017), and adopt fine-grained labels on the similarity between the input and output pairs, with a value of 5 indicating perfect match, down to 0 indicating unrelated pairs (See Table 1 of Cer et al. (2017) for the detailed description of the labels). We see that across languages, the input and output share good similarity (at least $50\% \geq 3$); however we do note that there is relatively larger amount of noise for *kn* and *ml* languages (with high values of Label 0).

We release the annotated data along with the datasets and the trained models.

## 5 Experimental Settings

We establish strong baseline models using pre-trained models and multilingualism. We describe the experimental settings for generating benchmark scores for all the tasks.

### 5.1 Datasets

We use the aforementioned datasets we created for our experiments and split them into 3 parts, roughly 80%, 10% and 10% for training, development and testing, respectively (with some exceptions). Details of splits are in the Appendix section B.5 and Appendix Tables 10, 13, 18 and 23.

### 5.2 Models Compared

We compare monolingual and multilingual fine-tuning of multilingual pre-trained models; strategies that are important for low-resource languages. By monolingual models, we mean models fine-tuned on data for one language. To specify the language, we prefix a language code for each example. A multilingual model is fine-tuned on the dataset obtained by combining all the languages' data.

### 5.3 Pre-trained Models Used

For our experiments with fine-tuning of pre-trained encoder-decoder Transformer (Vaswani et al., 2017) models. We compare IndicBART (Dabre et al., 2021), a *language-group specific pre-trained model* trained specifically for Indic languages, with mT5, a *general* pre-trained model for 100+ languages. We used mT5 instead of mBART (Liu et al., 2020) since it covers all languages in our dataset.

**IndicBART**: IndicBART (Dabre et al., 2021) is a pre-trained model that focuses on all 11 Indic languages in this paper, trained in the same way as mBART (Liu et al., 2020). It has two versions, one using the Devanagari script for all languages and another using each language's original scripts.
**mT5**: mT5 (Xue et al., 2021) is a multilingual model, trained using the span-prediction denoising approach, covering 101 languages, and we choose the mT5-small model containing 300M parameters for a fair comparison with the IndicBART model containing 244M parameters.

### 5.4 Implementations Used

We use YANMTT toolkit[10] (Dabre and Sumita, 2021) for fine-tuning IndicBART. For fine-tuning mT5, we use/modify the Huggingface (Wolf et al., 2020) scripts[11].

### 5.5 Training Settings

As much as possible, we tune hyperparameters when we fine-tune models, referring to settings in Dabre et al. (2021) and Xue et al. (2021). We give details about hyperparameter settings and training convergence in Appendix A.

---

### 5.6 Evaluation Metrics

For all tasks except paraphrasing, we report the Rouge-L F1 score (Lin, 2004). In order to compute Rouge scores for the decoded test sets, we use the multilingual rouge scoring implementation (Hasan et al., 2021)[12] which enables segmentation, stemming and punctuation normalization for various languages. For paraphrasing, we compute iBLEU (Sun and Zhou, 2012) following Hosking and Lapata (2021) using the equation: $iBLEU = \alpha * \text{BLEU}(O, R) - (1 - \alpha) * \text{BLEU}(O, I)$, where $O$ =output, $R$ =references, $I$ =input and, $\alpha = 0.7$. BLEU is calculated using sacreBLEU (Post, 2018). Higher iBLEU implies better paraphrasing.

## 6 Results and Analysis

In this section, we present the results obtained using models trained for a variety of tasks and analyze them from various perspectives. We compare between models fine-tuned on IndicBART (IB), separate script IndicBART (SSIB) and mT5 in monolingual and multilingual settings.

### 6.1 Research Questions

In the analysis presented below, we try to answer the following research questions:

**Impact of multilingualism:** How do monolingual models compare with multilingual models?

**Impact of language family:** Are language family specific pre-trained models (IndicBART) better than universal pre-trained models (mT5)?

**Impact of task nature on performance:** What are the determiners of the task performance? For this, we try to compare across tasks and languages and provide insights.

### 6.2 Main Results

We report the Rouge-L scores for all tasks in table 4: To save space we report on 5 languages: Assamese (as), Hindi (hi), Oriya (or), Tamil (ta) and Telugu (te), and give the rest in the Appendix which also show the impact of pre-training by comparing against models trained from scratch.

**Impact of multilingualism:** Multilingual models are inherently superior to monolingual models, regardless of training from scratch or via fine-tuning. This shows that multilingualism enables transfer learning, giving stronger baselines. The only ex-ception is headline generation. We suspect that this is due to poor hyperparameter choices.

| L | Monolingual | | | Multilingual | | |
|---|---|---|---|---|---|---|
| | mT5 | SSIB | IB | mT5 | SSIB | IB |
| **Biography Generation (Rouge-L)** | | | | | | |
| as | 49.53 | 55.21 | 55.68 | 56.48 | **56.50** | 56.28 |
| hi | 67.08 | 67.16 | 65.86 | **67.57** | 67.34 | 67.48 |
| or | 61.87 | 69.82 | 65.79 | 69.49 | **70.71** | 67.13 |
| ta | 51.60 | 51.14 | 51.69 | **52.36** | 51.11 | 51.82 |
| te | 51.53 | 50.89 | 50.25 | **52.22** | 51.72 | 51.43 |
| **Headline Generation (Rouge-L)** | | | | | | |
| as | 30.80 | 68.26 | **71.56** | 33.58 | 46.82 | 44.64 |
| hi | 32.55 | 34.49 | **34.57** | 32.68 | 33.60 | 32.70 |
| or | 21.22 | **23.93** | 21.95 | 21.94 | 23.74 | 21.62 |
| ta | 46.42 | 46.52 | **46.87** | 43.29 | 45.72 | 45.94 |
| te | 31.56 | 41.97 | **42.89** | 32.36 | 35.58 | 36.66 |
| **Sentence Summarization (Rouge-L)** | | | | | | |
| as | 42.58 | 62.65 | 60.13 | **70.26** | 62.57 | 59.29 |
| hi | 44.86 | 44.88 | **45.57** | 44.03 | 45.15 | 45.34 |
| or | 44.51 | 47.65 | 42.50 | 44.01 | **49.23** | 43.65 |
| ta | 56.64 | 56.60 | 56.16 | 55.85 | 56.49 | **56.83** |
| te | 47.48 | 52.33 | 52.62 | **53.65** | 52.66 | 53.44 |
| **Paraphrase Generation (iBLEU)** | | | | | | |
| as | 0.34 | 0.23 | 0.39 | 0.42 | 0.34 | **0.54** |
| hi | 8.58 | 17.29 | **18.55** | 9.48 | 17.01 | 18.24 |
| or | 0.35 | 14.2 | 10.33 | 1.24 | **15.31** | 12.85 |
| ta | 8.09 | 10.63 | 11.94 | 8.31 | 10.74 | **12.10** |
| te | 5.05 | 9.19 | **11.06** | 5.5 | 9.23 | 10.69 |
| **Question Generation (Rouge-L)** | | | | | | |
| as | 19.69 | 20.33 | 20.21 | **20.90** | 20.73 | 20.48 |
| hi | 34.58 | 33.60 | 32.24 | 34.14 | 34.42 | **35.38** |
| or | 20.34 | 25.70 | 24.29 | 20.90 | **27.53** | 25.25 |
| ta | 22.84 | 21.24 | 21.24 | 22.61 | **23.49** | 22.98 |
| te | 25.63 | 23.15 | 24.46 | 25.01 | **25.81** | 25.67 |

Table 4: iBLEU scores for paraphrasing and Rouge-L scores for biography generation (WikiBio), headline generation, sentence summarization and question generation. We report scores for 5 languages: Assamese (as), Hindi (hi), Oriya (or), Tamil (ta) and Telugu (te). We compare between models fine-tuned on mT5, separate script IndicBART (SSIB), single script IndicBART (IB) in monolingual and multilingual settings.

**Impact of language family:** In monolingual settings, with a few exceptions, fine-tuning IndicBART gives substantially better results than fine-tuning mT5. In multilingual settings, the gap narrows and in several cases, fine-tuned mT5 outperforms fine-tuned IndicBART. Note that mT5 contains 300M parameters, whereas IndicBART has 244M parameters. In monolingual settings, a language family specific pre-trained model seems to be more beneficial, in addition to being more cost-effective, as compared to a generic pre-trained model. However, in multilingual settings, a larger model might be better, regardless of its generic nature, as the additional parameters can be better utilized to learn from the increased data.

**Impact of task nature on performance:** Biogra-phy generation, headline generation and sentence summarization exhibit relatively higher Rouge-L scores. For biography and sentence summarization, the highest average scores are around 54 to 55 obtained by multilingual fine-tuning of mT5 and IndicBART. The best performance of headline generation is close to 49 obtained by monolingual fine-tuned IndicBART. Biography generation involves converting a table with disjoint information into a sentence and sentence summarization involves converting a sentence into a shorter one. Both tasks are relatively simpler, which may be the reason for the high scores. On the other hand, headline generation involves converting a document into a short sentence which is a relatively harder task, explaining why the scores are lower than for the other two tasks. For paraphrasing, multilingually fine-tuned IndicBART gives the best performance, but the iBLEU scores are less than 20. This shows that paraphrasing is a challenging task, as the model should learn to produce diverse paraphrases and not just make trivial changes to the in-put. Finally, question generation also proves to be a challenging task, as evidenced by the relatively lower Rouge scores close to 26.

| L | Rouge-L | | | | | |
|---|---|---|---|---|---|---|
| | Monolingual | | | Multilingual | | |
| | mT5 | SSIB | IB | mT5 | SSIB | IB |
| **C-hi** | 39.56 | 32.97 | 26.45 | **39.61** | 34.42 | 33.90 |
| **C-ta** | **26.49** | 21.23 | 17.76 | 23.36 | 21.02 | 17.94 |
| **M-hi** | 25.74 | 23.97 | 22.57 | **26.21** | 25.85 | 25.76 |
| **T-bn** | 27.23 | 20.02 | 20.32 | **27.02** | 24.04 | 22.46 |
| **T-te** | 30.17 | 22.93 | 23.24 | **30.91** | 27.67 | 24.68 |
| **X-hi** | 32.30 | 28.06 | 26.22 | **33.75** | 30.44 | 31.11 |
| **avg** | **30.25** | 24.86 | 22.76 | 30.14 | 27.24 | 25.98 |

Table 5: Rouge-L scores for question generation on different real world existing question generation datasets: chaii (C), MLQA (M), TyDi QA (T) and XQuAD (X). We compare between IndicBART (IB), separate script IndicBART (SSIB) and mT5 models in monolingual and multilingual settings.

| Task | Monolingual | | | Multilingual | | |
|---|---|---|---|---|---|---|
| | mT5 | SSIB | IB | mT5 | SSIB | IB |
| **BG** | 51.88 | 52.90 | 52.45 | **54.61** | 53.84 | 53.70 |
| **HG** | 38.43 | 48.08 | **48.60** | 45.46 | 42.45 | 43.71 |
| **SS** | 50.19 | 54.29 | 54.07 | **55.15** | 54.88 | 54.51 |
| **PG** | 4.60 | 9.88 | 10.22 | 5.11 | 10.06 | **10.64** |
| **QG** | 25.15 | 24.45 | 24.23 | 25.15 | **26.60** | 25.95 |

Table 6: Summary of results on IndicNLG Benchmark across biography generation (BG), headline generation (HG), sentence summarization (SS), paraphrase generation (PG) and question generation (QG). The table shows average scores across all 11 languages (iBLEU for paraphrase generation and Rouge-L for other tasks). We compare models fine-tuned on mT5, separate script IndicBART (SSIB) and single script IndicBART (IB) in monolingual and multilingual settings.

**Additional evaluation of QG models**: We also evaluate our fine-tuned QG models on publicly available manually created test sets (chaii (Google, 2021), MLQA (Lewis et al., 2020b), TyDi QA (Clark et al., 2020) and XQuAD (Artetxe et al., 2020) test sets (results in Table 5). Note that these test sets cover only 4 languages. We see that results on translated test sets and manual test sets indicate broadly similar relative ranking of models - indicating that translated test sets can serve as a reasonable proxy when manual test sets are not available. However, we note that score differences between mT5 and IndicBART models are significantly larger on the manual test sets.

### 6.3 Summary of IndicNLG Benchmarking

Table 6 gives an overview of average performance across all languages for each task in monolingual and multilingual settings. This helps us answer the research questions we posed earlier. Compared to the fine-grained view in Table 4, the observations still hold on an average, which shows that multilingual fine-tuning IndicBART models is highly useful regardless of language or task. Headline generation is an exception to this rule, but even in this task, IndicBART models achieves higher score. The overall competitiveness or superiority of IndicBART indicates the importance of language group specific pre-trained models for NLG. We also observe that task performance varies according to language as well as difficulty and according to scores alone, the perceived difficulty of tasks from easiest to hardest are: paraphrasing, question generation, headline generation, biography generation and sentence summarization.

### 6.4 Transfer Learning across tasks

We study if our models for one NLG task can benefit another task. In our case study, we explore whether our fine-tuned headline generation models can be further fine-tuned to improve extreme document summarization. Extreme summarization is the task of generating a short (often) one-sentence summary of a news article (Narayan et al., 2018; Hasan et al., 2021). Headline generation also compresses an article into a few words and can be seen as complementary task to extreme summarization. Specifically, for this study we focus on the Indic languages (*bn, gu, hi, mr, pa, ta,*

| L | Zero Shot | Supervised | |
| | IB → HG | IB → XL | IB → HG → XL |
|---|---|---|---|
| bn | 13.49 | 16.02 | **20.01** |
| gu | 11.34 | 13.40 | **17.20** |
| hi | 18.42 | 19.88 | **22.93** |
| mr | 11.93 | 16.04 | **23.04** |
| pa | 13.80 | 19.20 | **23.14** |
| ta | 13.41 | 16.63 | **22.01** |
| te | 11.08 | 14.42 | **19.48** |
| avg | 13.35 | 16.51 | **21.12** |

Table 7: XLSum test set evaluation on different experiments to show that headline generation dataset helps in a transfer learning setup. IB is IndicBART, XL is XL-Sum, HG is Headline Generation.

*te*) from the XL-Sum dataset (Hasan et al., 2021). Table 7 shows the results for (a) the zero-shot evaluation of multilingual headline generation models for XL-Sum summarization (IB → HG) and (b) supervised results using the XL-Sum training data for multilingual fine-tuning of IndicBART (IB → XL) and of headline generation models (IB → HG → XL). We see that its possible to obtain reasonable performance on summarization using a headline generation model as it is, although it is better to fine-tune IndicBART on the summarization data to get a boost of around 3 Rouge-L. However, fine-tuning the headline generation model significantly improves summarization (average 5 Rouge-L improvement) showing that our headline generation models may be used as pre-trained models for other down-stream tasks.

## 7 Conclusion and Future Work

We present the IndicNLG Benchmark, a collection of datasets for 5 diverse NLG tasks in 11 Indic languages and 8M examples, with the aim of creating much-required standard benchmarks to drive Indic NLG research. To the best of our knowledge, this is the most diverse multilingual NLG dataset. Our methods are simple enough to create similar datasets for modest resourced languages. We trained a variety of monolingual and multilingual models and showed the impact of the combination of multilingualism and pre-trained models. In general, multilingual models outperform their monolingual counterparts. Language-family specific pre-trained models like IndicBART are valuable as they are competitive with large multilingual models like mT5 despite needing only 14% the compute (147B vs 1T training tokens), while having smaller vocabularies (64K vs 250K) and fewer params (244M vs 300M). Given these ob-servations, we recommend future baselines to consider multilingual fine-tuning of language family-specific models as the starting point. Future work will also involve extending the datasets for additional Indic languages and new tasks.

## 8 Limitations

This paper describes methods for data creation for 11 Indic languages for the purposes of natural language generation along with modeling recommendations. The following are the limitations:

1. Data creation relies on resources like parallel corpora, monolingual corpora and Wikipedia of reasonable sizes. Our approach may not apply to languages where such resources are scarce.

2. For researchers with limited computational resources, training multilingual models may be time-consuming, especially given the sizes of datasets for paraphrasing and headline generation.

3. Evaluation for NLG is still not deeply explored and the metrics we use might not be the best ones, although they are widely used in existing literature.

4. The question generation dataset is generated using machine translation, and it might consist of translationese.

## Acknowledgements

We would like to thank the Ministry of Electronics and Information Technology (MeitY[13]) of the Government of India and the Centre for Development of Advanced Computing (C-DAC[14]), Pune for generously supporting this work and providing us access to multiple GPU nodes on the Param Siddhi Supercomputer. We would like to thank the Ek-Step Foundation and Nilekani Philanthropies for their generous grant which went into hiring human resources as well as cloud resources needed for this work. We would like to thank Anupama Sujatha from AI4Bharat for helping in coordinating the annotation task, and extend thanks to all the annotators of AI4Bharat team.

---

[13] https://www.meity.gov.in/
[14] https://www.cdac.in/index.aspx?id=pune

# References

Divyanshu Aggarwal, Vivek Gupta, and Anoop Kunchukuttan. 2022. Indicxnli: Evaluating multilingual inference for indian languages.

Husam Ali, Yllias Chali, and Sadid A. Hasan. 2010. Automatic question generation from sentences. In *Actes de la 17e conférence sur le Traitement Automatique des Langues Naturelles. Articles courts*, pages 213–218, Montréal, Canada. ATALA.

Mikel Artetxe, Sebastian Ruder, and Dani Yogatama. 2020. On the cross-lingual transferability of monolingual representations. In *Proceedings of the 58th Annual Meeting of the Association for Computational Linguistics*, pages 4623–4637, Online. Association for Computational Linguistics.

Michele Banko, Vibhu O. Mittal, and Michael J. Witbrock. 2000. Headline generation based on statistical translation. In *Proceedings of the 38th Annual Meeting of the Association for Computational Linguistics*, pages 318–325, Hong Kong. Association for Computational Linguistics.

Regina Barzilay and Lillian Lee. 2003. Learning to paraphrase: An unsupervised approach using multiple-sequence alignment. In *Proceedings of the 2003 Human Language Technology Conference of the North American Chapter of the Association for Computational Linguistics*, pages 16–23.

Emily Bender. 2019. The# benderrule: On naming the languages we study and why it matters. https://thegradient.pub/the-benderrule-on-naming-the-languages-we-study-and-why-it-matters. [Online; accessed 1-May-2022].

Jonathan Brown, Gwen Frishkoff, and Maxine Eskenazi. 2005. Automatic question generation for vocabulary assessment. In *Proceedings of Human Language Technology Conference and Conference on Empirical Methods in Natural Language Processing*, pages 819–826, Vancouver, British Columbia, Canada. Association for Computational Linguistics.

Samuel Cahyawijaya, Genta Indra Winata, Bryan Wilie, Karissa Vincentio, Xiaohong Li, Adhiguna Kuncoro, Sebastian Ruder, Zhi Yuan Lim, Syafri Bahar, Masayu Khodra, Ayu Purwarianti, and Pascale Fung. 2021. IndoNLG: Benchmark and resources for evaluating Indonesian natural language generation. In *Proceedings of the 2021 Conference on Empirical Methods in Natural Language Processing*, pages 8875–8898, Online and Punta Cana, Dominican Republic. Association for Computational Linguistics.

Daniel Cer, Mona Diab, Eneko Agirre, Iñigo Lopez-Gazpio, and Lucia Specia. 2017. SemEval-2017 task 1: Semantic textual similarity multilingual and crosslingual focused evaluation. In *Proceedings of the 11th International Workshop on Semantic Evaluation (SemEval-2017)*, pages 1–14, Vancouver, Canada. Association for Computational Linguistics.

Yiran Chen, Zhenqiao Song, Xianze Wu, Danqing Wang, Jingjing Xu, Jiaze Chen, Hao Zhou, and Lei Li. 2022. MTG: A Benchmarking Suite for Multilingual Text Generation. In *Findings of NAACL 2022 (to appear)*.

Sumit Chopra, Michael Auli, and Alexander M. Rush. 2016. Abstractive sentence summarization with attentive recurrent neural networks. In *Proceedings of the 2016 Conference of the North American Chapter of the Association for Computational Linguistics: Human Language Technologies*, pages 93–98, San Diego, California. Association for Computational Linguistics.

Jonathan H. Clark, Eunsol Choi, Michael Collins, Dan Garrette, Tom Kwiatkowski, Vitaly Nikolaev, and Jennimaria Palomaki. 2020. TyDi QA: A benchmark for information-seeking question answering in typologically diverse languages. *Transactions of the Association for Computational Linguistics*, 8:454–470.

Mathias Creutz. 2018. Open subtitles paraphrase corpus for six languages. In *Proceedings of the Eleventh International Conference on Language Resources and Evaluation (LREC 2018)*, Miyazaki, Japan. European Language Resources Association (ELRA).

Raj Dabre, Aizhan Imankulova, and Masahiro Kaneko. 2021. Studying the impact of document-level context on simultaneous neural machine translation. In *Proceedings of Machine Translation Summit XVIII: Research Track*, pages 202–214, Virtual. Association for Machine Translation in the Americas.

Raj Dabre, Himani Shrotriya, Anoop Kunchukuttan, Ratish Puduppully, Mitesh Khapra, and Pratyush Kumar. 2022. IndicBART: A pre-trained model for indic natural language generation. In *Findings of the Association for Computational Linguistics: ACL 2022*, pages 1849–1863, Dublin, Ireland. Association for Computational Linguistics.

Raj Dabre and Eiichiro Sumita. 2021. YANMTT: yet another neural machine translation toolkit. *CoRR*, abs/2108.11126.

Li Dong, Nan Yang, Wenhui Wang, Furu Wei, Xiaodong Liu, Yu Wang, Jianfeng Gao, Ming Zhou, and Hsiao-Wuen Hon. 2019. Unified language model pre-training for natural language understanding and generation. *Advances in Neural Information Processing Systems*, 32.

Xinya Du, Junru Shao, and Claire Cardie. 2017. Learning to ask: Neural question generation for reading comprehension. In *Proceedings of the 55th Annual Meeting of the Association for Computational Linguistics (Volume 1: Long Papers)*, pages 1342–1352, Vancouver, Canada. Association for Computational Linguistics.

Ondřej Dušek, Jekaterina Novikova, and Verena Rieser. 2020. Evaluating the state-of-the-art of end-to-end natural language generation: The e2e nlg challenge. *Computer Speech & Language*, 59:123–156.

M. B. Emeneau. 1956. India as a linguistic area. *Language*, 32(1):3–16.

Claire Gardent, Anastasia Shimorina, Shashi Narayan, and Laura Perez-Beltrachini. 2017. Creating training corpora for NLG micro-planners. In *Proceedings of the 55th Annual Meeting of the Association for Computational Linguistics (Volume 1: Long Papers)*, pages 179–188, Vancouver, Canada. Association for Computational Linguistics.

Albert Gatt and Emiel Krahmer. 2018. Survey of the state of the art in natural language generation: Core tasks, applications and evaluation. *J. Artif. Int. Res.*, 61(1):65170.

Sebastian Gehrmann, Tosin Adewumi, Karmanya Aggarwal, Pawan Sasanka Ammanamanchi, Anuoluwapo Aremu, Antoine Bosselut, Khyathi Raghavi Chandu, Miruna-Adriana Clinciu, Dipanjan Das, Kaustubh Dhole, Wanyu Du, Esin Durmus, Ondřej Dušek, Chris Chinenye Emezue, Varun Gangal, Cristina Garbacea, Tatsunori Hashimoto, Yufang Hou, Yacine Jernite, Harsh Jhamtani, Yangfeng Ji, Shailza Jolly, Mihir Kale, Dhruv Kumar, Faisal Ladhak, Aman Madaan, Mounica Maddela, Khyati Mahamood, Saad Mahamood, Bodhisattwa Prasad Majumder, Pedro Henrique Martins, Angelina McMillan-Major, Simon Mille, Emiel van Miltenburg, Moin Nadeem, Shashi Narayan, Vitaly Nikolaev, Andre Niyongabo Rubungo, Salomey Osei, Ankur Parikh, Laura Perez-Beltrachini, Niranjan Ramesh Rao, Vikas Raunak, Juan Diego Rodriguez, Sashank Santhanam, João Sedoc, Thibault Sellam, Samira Shaikh, Anastasia Shimorina, Marco Antonio Sobrevilla Cabezudo, Hendrik Strobelt, Nishant Subramani, Wei Xu, Diyi Yang, Akhila Yerukola, and Jiawei Zhou. 2021. The GEM benchmark: Natural language generation, its evaluation and metrics. In *Proceedings of the 1st Workshop on Natural Language Generation, Evaluation, and Metrics (GEM 2021)*, pages 96–120, Online. Association for Computational Linguistics.

Google. 2021. chaii - Hindi and Tamil question answering. https://www.kaggle.com/c/chaii-hindi-and-tamil-question-answering. [Online; accessed 1-May-2022].

Jian Guan, Zhuoer Feng, Yamei Chen, Ruilin He, Xiaoxi Mao, Changjie Fan, and Minlie Huang. 2022. LOT: A story-centric benchmark for evaluating Chinese long text understanding and generation. *Transactions of the Association for Computational Linguistics*, 10:434–451.

Tahmid Hasan, Abhik Bhattacharjee, Md. Saiful Islam, Kazi Mubasshir, Yuan-Fang Li, Yong-Bin Kang, M. Sohel Rahman, and Rifat Shahriyar. 2021. XL-sum: Large-scale multilingual abstractive summarization for 44 languages. In *Findings of the Association for Computational Linguistics: ACL-IJCNLP 2021*, pages 4693–4703, Online. Association for Computational Linguistics.

Michael Heilman. 2011. *Automatic Factual Question Generation from Text*. Ph.D. thesis, Carnegie Mellon University, USA. AAI3528179.

Tom Hosking and Mirella Lapata. 2021. Factorising meaning and form for intent-preserving paraphrasing. In *Proceedings of the 59th Annual Meeting of the Association for Computational Linguistics and the 11th International Joint Conference on Natural Language Processing (Volume 1: Long Papers)*, pages 1405–1418, Online. Association for Computational Linguistics.

Divyanshu Kakwani, Anoop Kunchukuttan, Satish Golla, Gokul N.C., Avik Bhattacharyya, Mitesh M. Khapra, and Pratyush Kumar. 2020. IndicNLPSuite: Monolingual corpora, evaluation benchmarks and pre-trained multilingual language models for Indian languages. In *Findings of the Association for Computational Linguistics: EMNLP 2020*, pages 4948–4961, Online. Association for Computational Linguistics.

Mihir Kale and Abhinav Rastogi. 2020. Text-to-text pre-training for data-to-text tasks. In *Proceedings of the 13th International Conference on Natural Language Generation*, pages 97–102, Dublin, Ireland. Association for Computational Linguistics.

Anoop Kunchukuttan. 2020. The IndicNLP Library. https://github.com/anoopkunchukuttan/indic_nlp_library/blob/master/docs/indicnlp.pdf.

Faisal Ladhak, Esin Durmus, Claire Cardie, and Kathleen McKeown. 2020. WikiLingua: A new benchmark dataset for cross-lingual abstractive summarization. In *Findings of the Association for Computational Linguistics: EMNLP 2020*, pages 4034–4048, Online. Association for Computational Linguistics.

Rémi Lebret, David Grangier, and Michael Auli. 2016. Neural text generation from structured data with application to the biography domain. In *Proceedings of the 2016 Conference on Empirical Methods in Natural Language Processing*, pages 1203–1213, Austin, Texas. Association for Computational Linguistics.

Mike Lewis, Yinhan Liu, Naman Goyal, Marjan Ghazvininejad, Abdelrahman Mohamed, Omer Levy, Veselin Stoyanov, and Luke Zettlemoyer. 2020a. BART: Denoising sequence-to-sequence pre-training for natural language generation, translation, and comprehension. In *Proceedings of the 58th Annual Meeting of the Association for Computational Linguistics*, pages 7871–7880, Online. Association for Computational Linguistics.

Patrick Lewis, Barlas Oguz, Ruty Rinott, Sebastian Riedel, and Holger Schwenk. 2020b. MLQA: Evaluating cross-lingual extractive question answering. In *Proceedings of the 58th Annual Meeting of the Association for Computational Linguistics*, pages 7315–7330, Online. Association for Computational Linguistics.

Chin-Yew Lin. 2004. ROUGE: A package for automatic evaluation of summaries. In *Text Summarization Branches Out*, pages 74–81, Barcelona, Spain. Association for Computational Linguistics.

Yinhan Liu, Jiatao Gu, Naman Goyal, Xian Li, Sergey Edunov, Marjan Ghazvininejad, Mike Lewis, and Luke Zettlemoyer. 2020. Multilingual denoising pre-training for neural machine translation. *Transactions of the Association for Computational Linguistics*, 8:726–742.

Nitin Madnani and Bonnie J. Dorr. 2010. Generating phrasal and sentential paraphrases: A survey of data-driven methods. *Computational Linguistics*, 36(3):341–387.

Kathleen R. McKeown. 1979. Paraphrasing using given and new information in a question-answer system. In *17th Annual Meeting of the Association for Computational Linguistics*, pages 67–72, La Jolla, California, USA. Association for Computational Linguistics.

Kathleen R. McKeown. 1983. Paraphrasing questions using given and new information. *American Journal of Computational Linguistics*, 9(1):1–10.

Marie Meteer and Varda Shaked. 1988. Strategies for effective paraphrasing. In *Coling Budapest 1988 Volume 2: International Conference on Computational Linguistics*.

Linyong Nan, Dragomir Radev, Rui Zhang, Amrit Rau, Abhinand Sivaprasad, Chiachun Hsieh, Xiangru Tang, Aadit Vyas, Neha Verma, Pranav Krishna, Yangxiaokang Liu, Nadia Irwanto, Jessica Pan, Faiaz Rahman, Ahmad Zaidi, Mutethia Mutuma, Yasin Tarabar, Ankit Gupta, Tao Yu, Yi Chern Tan, Xi Victoria Lin, Caiming Xiong, Richard Socher, and Nazneen Fatema Rajani. 2021. DART: Open-domain structured data record to text generation. In *Proceedings of the 2021 Conference of the North American Chapter of the Association for Computational Linguistics: Human Language Technologies*, pages 432–447, Online. Association for Computational Linguistics.

Shashi Narayan, Shay B. Cohen, and Mirella Lapata. 2018. Don't give me the details, just the summary! topic-aware convolutional neural networks for extreme summarization. In *Proceedings of the 2018 Conference on Empirical Methods in Natural Language Processing*, pages 1797–1807, Brussels, Belgium. Association for Computational Linguistics.

Ankur Parikh, Xuezhi Wang, Sebastian Gehrmann, Manaal Faruqui, Bhuwan Dhingra, Diyi Yang, and Dipanjan Das. 2020. ToTTo: A controlled table-to-text generation dataset. In *Proceedings of the 2020 Conference on Empirical Methods in Natural Language Processing (EMNLP)*, pages 1173–1186, Online. Association for Computational Linguistics.

Matt Post. 2018. A call for clarity in reporting BLEU scores. In *Proceedings of the Third Conference on Machine Translation: Research Papers*, pages 186–191, Brussels, Belgium. Association for Computational Linguistics.

Ratish Puduppully and Mirella Lapata. 2021. Data-to-text generation with macro planning. *Transactions of the Association for Computational Linguistics*, 9:510–527.

Colin Raffel, Noam Shazeer, Adam Roberts, Katherine Lee, Sharan Narang, Michael Matena, Yanqi Zhou, Wei Li, and Peter J. Liu. 2020. Exploring the limits of transfer learning with a unified text-to-text transformer. *Journal of Machine Learning Research*, 21(140):1–67.

Pranav Rajpurkar, Jian Zhang, Konstantin Lopyrev, and Percy Liang. 2016. SQuAD: 100,000+ questions for machine comprehension of text. In *Proceedings of the 2016 Conference on Empirical Methods in Natural Language Processing*, pages 2383–2392, Austin, Texas. Association for Computational Linguistics.

Gowtham Ramesh, Sumanth Doddapaneni, Aravinth Bheemaraj, Mayank Jobanputra, Raghavan AK, Ajitesh Sharma, Sujit Sahoo, Harshita Diddee, Mahalakshmi J, Divyanshu Kakwani, Navneet Kumar, Aswin Pradeep, Srihari Nagaraj, Kumar Deepak, Vivek Raghavan, Anoop Kunchukuttan, Pratyush Kumar, and Mitesh Shantadevi Khapra. 2022. Samanantar: The Largest Publicly Available Parallel Corpora Collection for 11 Indic Languages. *Transactions of the Association for Computational Linguistics*, 10:145–162.

Ehud Reiter and Robert Dale. 1997. Building applied natural language generation systems. *Nat. Lang. Eng.*, 3(1):57–87.

Alexander M. Rush, Sumit Chopra, and Jason Weston. 2015. A neural attention model for abstractive sentence summarization. In *Proceedings of the 2015 Conference on Empirical Methods in Natural Language Processing*, pages 379–389, Lisbon, Portugal. Association for Computational Linguistics.

Thomas Scialom, Paul-Alexis Dray, Sylvain Lamprier, Benjamin Piwowarski, and Jacopo Staiano. 2020. MLSUM: The multilingual summarization corpus. In *Proceedings of the 2020 Conference on Empirical Methods in Natural Language Processing (EMNLP)*, pages 8051–8067, Online. Association for Computational Linguistics.

Hong Sun and Ming Zhou. 2012. Joint learning of a dual SMT system for paraphrase generation. In *Proceedings of the 50th Annual Meeting of the Association for Computational Linguistics (Volume 2: Short Papers)*, pages 38–42, Jeju Island, Korea. Association for Computational Linguistics.

Yuqing Tang, Chau Tran, Xian Li, Peng-Jen Chen, Naman Goyal, Vishrav Chaudhary, Jiatao Gu, and Angela Fan. 2021. Multilingual translation from denoising pre-training. In *Findings of the Association for Computational Linguistics: ACL-IJCNLP 2021*, pages 3450–3466, Online. Association for Computational Linguistics.

Ashish Vaswani, Noam Shazeer, Niki Parmar, Jakob Uszkoreit, Llion Jones, Aidan N Gomez, Ł ukasz Kaiser, and Illia Polosukhin. 2017. Attention is all you need. In I. Guyon, U. V. Luxburg, S. Bengio, H. Wallach, R. Fergus, S. Vishwanathan, and R. Garnett, editors, *Proceedings of the Advances in Neural Information Processing Systems 30*, pages 5998–6008. Curran Associates, Inc.

Sam Wiseman, Stuart Shieber, and Alexander Rush. 2017. Challenges in data-to-document generation. In *Proceedings of the 2017 Conference on Empirical Methods in Natural Language Processing*, pages 2253–2263, Copenhagen, Denmark. Association for Computational Linguistics.

Thomas Wolf, Lysandre Debut, Victor Sanh, Julien Chaumond, Clement Delangue, Anthony Moi, Pierric Cistac, Tim Rault, Remi Louf, Morgan Funtowicz, Joe Davison, Sam Shleifer, Patrick von Platen, Clara Ma, Yacine Jernite, Julien Plu, Canwen Xu, Teven Le Scao, Sylvain Gugger, Mariama Drame, Quentin Lhoest, and Alexander Rush. 2020. Transformers: State-of-the-art natural language processing. In *Proceedings of the 2020 Conference on Empirical Methods in Natural Language Processing: System Demonstrations*, pages 38–45, Online. Association for Computational Linguistics.

Linting Xue, Noah Constant, Adam Roberts, Mihir Kale, Rami Al-Rfou, Aditya Siddhant, Aditya Barua, and Colin Raffel. 2021. mT5: A massively multilingual pre-trained text-to-text transformer. In *Proceedings of the 2021 Conference of the North American Chapter of the Association for Computational Linguistics: Human Language Technologies*, pages 483–498, Online. Association for Computational Linguistics.

Shiqi Zhao, Haifeng Wang, Ting Liu, and Sheng Li. 2008. Pivot approach for extracting paraphrase patterns from bilingual corpora. In *Proceedings of ACL-08: HLT*, pages 780–788, Columbus, Ohio. Association for Computational Linguistics.

Qingyu Zhou, Nan Yang, Furu Wei, Chuanqi Tan, Hangbo Bao, and M. Zhou. 2017. Neural question generation from text: A preliminary study. In *NLPCC*.

## A  Hyperparameter Tuning

We provide details of model hyperparamers we used for model training. All our models were fine-tuned on single A100 GPUs, regardless of monolingual or multilingual fine-tuning. For fine-tuning IndicBART, we followed the settings recommended by Dabre et al. (2022) and for fine-tuning we mT5-small we followed those recommended by Xue et al. (2021). For IndicBART we train till convergence on the validation set scores, which are computed via greedy decoding every 1,000 batches. On the other hand, we train mT5 for 10 epochs and choose the checkpoint with the highest validation set scores. In case of multilingual models, the convergence is determined by the average of validation set scores for all languages. For decoding the test sets, we use beam search. Following are some specific optimal hyperparameters for each task which we determined to be optimal:

**Biography Generation (WikiBio)**: For IndicBART fine-tuning we use batch sizes of 4096 tokens, dropouts of 0.1, label smoothing of 0.1, learning rate of 0.0003 and weight decay of 0.00001 with the ADAM optimizer. For mT5, we used most of the default hyperparameters in the fine-tuning script, with exceptions: batch sizes of 32 examples and learning rate of 0.00005 with the ADAM optimizer. We use maximum source and target length of 512 and 64 respectively for both models. For decoding the test sets, we use beam search with a beam of size 4, length penalty of 1.0 and penalize translations in the beam where 4-grams are repeated.

**Headline Generation**: Settings for IndicBART are the same as for WikiBio. We train monolingual mT5 models for 10 epochs with learning rate 5e-4 and weight decay 0.01. Multilingual mT5 models were trained for 15 epochs as our headline generation data is very large. For decoding the test sets, we use beam search with a beam of size 5, length penalty of 1.0 and penalize translations in the beam where 4-grams are repeated.

**Sentence Summarization**: The settings are the same as in WikiBio. For decoding the test sets, we use beam search with a beam of size 5, length penalty of 0.8 and penalize translations in the beam where 3-grams are repeated.

**Paraphrase Generation**: We use the same settings as for WikiBio. For decoding the test sets, we use beam search with a beam of size 4, length penalty of 1.0, and we do not use repetition penalties.

**Question Generation**: Other than using maximum source and target length of 256 and 32 respectively, we used the same settings as for WikiBio. For decoding the test sets, we use beam search with a beam of size 4, length penalty of 1.0, and we do not use repetition penalties.

## B  Dataset and Experiment Details

In the Appendices B.1 to B.5, we provide additional details on creation of the datasets, various dataset statistics, detailed results for all tasks and sample outputs.

### B.1  WikiBio

**Dataset Creation** The dataset creation process involves collection of raw data, noise removal, serialization, filtration, cross-lingual overlap removal and splitting.

**Raw Data Collection:** We download the Wikipedia[15] dumps for each language, parse them and save the information (metadata) of pages that have infoboxes. Next, we use the wptools API[16] to extract the infoboxes and the first lines of the pages about people. The first sentence is supposed to be a simple biography of the person whom the page is about.

| Field | Value | Transliteration |
|---|---|---|
| name | अमिताभ बच्चन | Amitabh Bachchan |
| spouse | जया बच्चन | Jaya Bachchan |
| father | हरिवंश राय बच्चन | Harivansh Rai Bachchan |

Table 8: A WikiBio infobox snippet for the Indian actor **Amitabh Bachchan**. The complete infobox will contain several facts as key-value pairs.

**Dataset Preprocessing:** The extracted data is rather unclean, as it contains spurious newline characters[17], special characters[18], and the values of some fields of the infobox are inside double square brackets([[ ]]). This necessitates cleaning

---

of the infoboxes and the single sentences we extract. Following cleaning, we serialize the infoboxes to convert it into a text sequence. Thus, we transform data-to-text generation into a text-to-text generation setup (Kale and Rastogi, 2020; Puduppully and Lapata, 2021). We separate attribute names from the values by enclosing the attributes names within special tokens. For example, the infobox snippet represented in Table 8 (excluding the transliteration column) gets converted into "<TAG> name </TAG> अमिताभ बच्चन <TAG> spouse </TAG>जया बच्चन <TAG> father </TAG> हरिवंश राय बच्चन" where "<TAG>" and "</TAG>" are tokens indicating that the content enclosed in them are fields.

We perform spelling normalization of the words in the infoboxes and the sentences describing them. Normalizing text written in Indic scripts helps to handle texts that display a lot of quirky behavior on account of varying input methods, multiple representations for the same character, etc. There is a need to canonicalize text representation so that NLP applications can consistently handle the data.

**Dataset cleaning and splitting:** We clean the dataset by discarding examples which satisfy the following criteria: output sentence containing fewer than 5 tokens, name field is not present in the input infobox, duplicate examples, and person name is in English. Furthermore, we clean the dataset to ensure there is no leakage during training of models in a multilingual setting. Otherwise, an example in one language in the training set may have its equivalent in another language in the validation or test set.

**Dataset Statistics** The dataset splits for each language are given in Table 9. We compare the counts of examples with that of English WikiBio dataset (Lebret et al., 2016). We see that the Indic language WikiBio is low resource as compared to English WikiBio, with total number of examples ∼6% of the size of English WikiBio.

In Table 10, we present some quantitative statistics where, we can see that the average count of words in input, output, count of attribute-value pairs, common words, and overlap percentage are comparable between Indic and English WikiBio.

**Results**: We report the rouge-L scores on the test set for all the 9 languages in Table 11.

**Example Outputs**: We present an example and its output generated by all the different models in table 12.

| Languages | Train | Test | Validation |
|-----------|-------|------|------------|
| **en** | 582,659 | 72,831 | 72,831 |
| **as** | 1,300 | 391 | 381 |
| **bn** | 4,615 | 1,521 | 1,567 |
| **hi** | 5,684 | 1,919 | 1,853 |
| **kn** | 1,188 | 389 | 383 |
| **ml** | 5,620 | 1,835 | 1,896 |
| **or** | 1,687 | 558 | 515 |
| **pa** | 3,796 | 1,227 | 1,331 |
| **ta** | 8,169 | 2,701 | 2,632 |
| **te** | 2,594 | 854 | 820 |
| **total** | 34,653 | 11,395 | 11,378 |

Table 9: Number of examples in the WikiBio dataset for 9 Indian languages. The total number of examples (last row) does not include examples in English, the statistics for which are obtained from Lebret et al. (2016).

| Languages | $W_{avg}^I$ | $W_{avg}^O$ | $W_{avg}^{FV}$ | $W_{avg}^{Common}$ | % overlap |
|-----------|-------------|-------------|----------------|--------------------|-----------|
| **en** | 112.06 | 26.06 | 19.67 | 11.57 | 44.40 |
| **as** | 144.48 | 16.50 | 13.27 | 4.82 | 29.21 |
| **bn** | 178.24 | 18.94 | 16.95 | 5.87 | 30.99 |
| **hi** | 146.50 | 17.46 | 17.62 | 6.29 | 36.03 |
| **kn** | 136.30 | 13.87 | 13.83 | 3.73 | 26.89 |
| **ml** | 148.89 | 14.24 | 13.83 | 3.47 | 24.37 |
| **or** | 131.89 | 14.31 | 14.02 | 4.14 | 28.93 |
| **pa** | 146.74 | 20.57 | 13.84 | 6.30 | 30.63 |
| **ta** | 163.55 | 16.92 | 19.79 | 5.32 | 31.44 |
| **te** | 119.78 | 11.69 | 12.75 | 3.69 | 31.57 |
| **average** | 132.13 | 14.10 | 13.01 | 4.26 | 30.21 |

Table 10: Quantitative statistics of the WikiBio dataset for English and 9 Indian languages. We use statistics such as the average number of words in input ($W_{avg}^I$), the average number of words in output ($W_{avg}^O$), the average number of field-value pairs ($W_{avg}^{FV}$), the average number of words common between input and output ($W_{avg}^{Common}$), and the percentage of words common between input and output (% overlap). Here, average statistics (last row) does not include English.

| | Rouge-L | | | | | | | |
|---|---|---|---|---|---|---|---|---|
| **Language** | **Monolingual** | | | | **Multilingual** | | | |
| | No PT | mT5 | SSIB | IB | No PT | mT5 | SSIB | IB |
| **as** | 13.67 | 49.53 | 55.21 | 55.68 | 49.16 | 56.48 | **56.50** | 56.28 |
| **bn** | 15.32 | 57.51 | 56.83 | 56.84 | 51.90 | **57.99** | 56.58 | 57.42 |
| **hi** | 57.49 | 68.07 | 67.16 | 65.86 | 62.79 | **67.57** | 67.34 | 67.48 |
| **kn** | 4.57 | 36.13 | 37.27 | 38.46 | 33.04 | **42.47** | 39.37 | 40.01 |
| **ml** | 6.05 | 38.79 | 37.42 | 37.95 | 33.34 | **39.60** | 38.42 | 38.84 |
| **or** | 23.33 | 61.87 | 69.82 | 65.79 | 62.69 | 69.49 | **70.71** | 67.13 |
| **pa** | 15.69 | 52.85 | 50.40 | 49.57 | 46.40 | **53.33** | 52.78 | 52.88 |
| **ta** | 44.14 | 51.60 | 51.14 | 51.69 | 47.26 | **52.36** | 51.11 | 51.82 |
| **te** | 10.39 | 51.53 | 50.89 | 50.25 | 43.79 | **52.22** | 51.72 | 51.43 |

Table 11: The table shows Rouge-L scores for the WikiBio test sets. We compare between models without pre-training (no PT), Indic BART (IB), separate script Indic BART (SSIB) and mT5 models in monolingual and multilingual settings.

| | | |
|---|---|---|
| **Input** | | \<TAG\> name \</TAG\> दुलू महतो \<TAG\> office \</TAG\> विधायक - बाघमारा विधानसभा निर्वाचन क्षेत्र - झारखंड । बाघमारा , झारखण्ड \<TAG\> term \</TAG\> दिसंबर 2014 से दिसंबर 2019 \<TAG\> nationality \</TAG\> भारतीय<br>**Translation**: \<TAG\> name \</TAG\> Dulu Mahto \<TAG\> office \</TAG\> MLA - Baghmara Assembly Constituency - Jharkhand. Baghmara , Jharkhand \<TAG\> term \</TAG\> December 2014 to December 2019 \<TAG\> nationality \</TAG\> Indian<br>**Transliteration**: \<tag\> name \</tag\> Dulu mahato \<tag\> office \</tag\> vidhaayak - baaghaamaara vidhaanasabha nirvaachan kshetra - jharkhand . baaghamaara , jharkhand \<tag\> term \</tag\> disambar 2014 se disambar 2019 \<tag\> nationality \</tag\> bharatiya |
| **Target Output** | | दुलू महतो भारत के झारखण्ड राज्य की बाघमारा सीट से भारतीय जनता पार्टी के विधायक हैं।<br>**Translation:** Dulu Mahto is a Bharatiya Janata Party MLA from Baghmara seat in the state of Jharkhand, India.<br>**Transliteration**: Dulu Mahto bhaarat ke jhrkhand raajya kee baaghamaara seat se bhaaratiya janata party ke vidhaayak hain. |
| **Monolingual** | **IB** | दुलू महतो भारत के झारखण्ड राज्य की बाघमारा सीट से झारखण्ड मुक्ति मोर्चा के विधायक हैं।<br>**Translation**: Dulu Mahto is the MLA of Jharkhand Mukti Morcha from Baghmara seat of Jharkhand state, India.<br>**Transliteration**: duloo mahato bhaarat ke jhaarakhand raajy kee baaghamaara seet se jhaarakhand mukti morcha ke vidhaayak hain. |
| | **No PT** | दुलू महतो एक भारतीय राजनीतिज्ञ तथा वर्तमान में झारखंड विधानसभा में विधायक हैं।<br>**Translation**: Dulu Mahto is an Indian politician and currently an MLA in the Jharkhand Legislative Assembly.<br>**Transliteration**: duloo mahato ek bhaarateey raajaneetigy tatha vartamaan mein jhaarakhand vidhaanasabha mein vidhaayak hain. |
| | **SSIB** | दुलू महतो भारत के झारखण्ड राज्य की बाघमारा सीट से झारखण्ड मुक्ति मोर्चा के विधायक हैं।<br>**Translation**:Dulu Mahto is the MLA of Jharkhand Mukti Morcha from Baghmara seat of Jharkhand state, India.<br>**Transliteration**: duloo mahato bhaarat ke jhaarakhand raajy kee baaghamaara seet se jhaarakhand mukti morcha ke vidhaayak hain. |
| | **mT5** | दुलू महतो भारत के झारखण्ड राज्य की बाघमारा सीट से भारतीय जनता पार्टी के विधायक हैं।<br>**Translation**: Dulu Mahto is a Bharatiya Janata Party MLA from Baghmara seat in the state of Jharkhand, India.<br>**Transliteration**: duloo mahato bhaarat ke jhaarakhand raajy kee baaghamaara seet se bhaarateey janata paartee ke vidhaayak hain. |
| **Multilingual** | **IB** | दुलू महतो भारत के झारखण्ड राज्य की बाघमारा सीट से भारतीय जनता पार्टी के विधायक हैं।<br>**Translation**: Dulu Mahto is a Bharatiya Janata Party MLA from Baghmara seat in the state of Jharkhand, India.<br>**Transliteration**: duloo mahato bhaarat ke jhaarakhand raajy kee baaghamaara seet se bhaarateey janata paartee ke vidhaayak hain. |
| | **No PT** | दुलू महतो भारत के झारखण्ड राज्य की बाघमारा सीट से भारतीय जनता पार्टी के विधायक हैं।<br>**Translation**:Dulu Mahto is a Bharatiya Janata Party MLA from Baghmara seat in the state of Jharkhand, India.<br>**Transliteration**: duloo mahato bhaarat ke jhaarakhand raajy kee baaghamaara seet se bhaarateey janata paartee ke vidhaayak hain. |
| | **SSIB** | दुलू महतो भारत के झारखण्ड राज्य की बाघमारा सीट से भारतीय जनता पार्टी के विधायक हैं।<br>**Translation**: Dulu Mahto is a Bharatiya Janata Party MLA from Baghmara seat in the state of Jharkhand, India.<br>**Transliteration**: duloo mahato bhaarat ke jhaarakhand raajy kee baaghamaara seet se bhaarateey janata paartee ke vidhaayak hain. |
| | **mT5** | दुलू महतो भारत के झारखण्ड राज्य की बाघमारा सीट से भारतीय जनता पार्टी के विधायक हैं।<br>**Translation**: Dulu Mahto is a Bharatiya Janata Party MLA from Baghmara seat in the state of Jharkhand, India.<br>**Transliteration**: duloo mahato bhaarat ke jhaarakhand raajy kee baaghamaara seet se bhaarateey janata paartee ke vidhaayak hain. |

Table 12: Model generated output for WikiBio

## B.2 Headline Generation

**Dataset Creation:** As mentioned in section 3.2.2, the raw data for the Hindi dataset is crawled from various web pages from different domains. Whereas for other languages, we use IndicGLUE headline classification dataset. The extraction of the article-title pair for IndicGLUE dataset is implicit, the correct headline is the title, and the article is the same. But for the Hindi dataset, we use manual inspection of these domains to get the logic for extracting the article-title pair. The first sentence of the body field is the title in some domains, whereas somewhere in the middle of the body field for others. Similarly, news articles are present in these body fields of the crawling. Hence it is an involved step to extricate the actual dataset from these crawlings.

**Dataset Cleaning:** The data has noises like publisher name and information, data and update time, or some advertisement links, or read more links. To clean the data majorly we perform regex pattern matching and keyword searching to find and remove the unwanted noise in the data. Sometimes a sample contains more than one language, or some domains has news in multiple languages. Hence to separate these languages we use language detection tools like gcld3[19], langdetect[20] and, langid[21]. The dataset created using the above process sometimes contains noise in the form of a document matched with an incorrect headline. We notice that such examples have low percentage of words in common between title and document. In order to remove such examples, we compute the *overlap* between the document content and the title by an "overlapping ratio". Suppose $D$ and $T$ represent the set of words in the document and title, respectively. The overlapping ratio is computed as $\frac{|D \cap T|}{|T|}$. We exclude examples below certain threshold of overlapping ratio.

**Dataset statistics:**

Table 13 gives the dataset splits. We calculate two types of dataset statistics, quantitative and qualitative.

**Quantitative Analysis:** Table 14 shows the percentage of novel n-grams, which indicates the percentage of n-grams of title not present in the document. It is a measure of the abstractive nature of the task, as the model will be required to predict words not present in the input. We find that the Indic Headline Generation dataset is comparable with XL-Sum dataset in terms of novel n-grams.

**Qualitative Analysis:** We use the LEAD-1 and EXT-ORACLE Rouge-L (R-L) scores as baselines, which also serve as an indicator of task difficulty. LEAD-1 Rouge-L scores are calculated between the first sentence of the document as system summary and title as reference summary. EXT-ORACLE scores are computed by selecting the sentence from the document as summary that give the highest rouge scores with the reference summary. Table 15 shows the LEAD-1 Rouge-L scores are very low for all the languages including XL-sum, indicating that the first sentence does not contain sufficient information to make a title. On the other hand, the EXT-ORACLE scores are higher. The scores clearly indicate that the task of headline generation is not a trivial one. Note that, the LEAD-1 and EXT-ORACLE scores in the case of document-headline pairs from XL-sum are substantially lower than our datasets, despite the low morphological complexity of English.

**Results:** Table 16 shows the Rouge-L scores for the headline generation test set across all the eleven languages and eight models. Monolingual IB gives the highest rouge-L score for almost all the languages.

**Output:** Table 17 shows the output generated by each model we have trained. Along with the native language of the example (which is Hindi here), we show the text's translation and transliteration in English for better understanding. We take only the first few sentences for the input, as the actual input article size is more than ten sentences. The multilingual IB and SSIB give a title which relates more to the article's first sentence. In contrast, monolingual IB and SSIB provide the overall summary in one line as the title, which correlates with the target summary. No PT in both monolingual and multilingual settings behaves the same, except multilingual output has more details.

| Dataset | Train | Test | Validation |
|---|---|---|---|
| **XL-Sum En** | 306,522 | 11,535 | 11,535 |
| **as** | 29,631 | 14,808 | 14,592 |
| **bn** | 113,424 | 14,568 | 14,739 |
| **gu** | 199,972 | 31,215 | 31,270 |
| **hi** | 208,091 | 44,475 | 44,718 |
| **kn** | 132,380 | 3,261 | 19,416 |
| **ml** | 10,358 | 5,220 | 5,388 |
| **mr** | 114,000 | 14,340 | 14,250 |
| **or** | 58,225 | 7,137 | 7,484 |
| **pa** | 48,441 | 6,086 | 6,108 |
| **ta** | 60,650 | 7,688 | 7,616 |
| **te** | 21,352 | 2,675 | 2,690 |
| **total** | 996,524 | 151,473 | 168,271 |

Table 13: Train, test and validation set sizes for headline generation in terms of number of samples. The 'total' row indicates the sum of the respective sets for all the languages.

| Dataset | avg. document length | | avg. title length | | vocabulary size | |
|---|---|---|---|---|---|---|
| | words | sentences | words | sentences | document | title |
| **XL-Sum En** | 460.42 | 24.38 | 8.15 | 1.06 | 1,465,523 | 146,683 |
| **as** | 185.70 | 8.37 | 7.94 | 1.01 | 246,387 | 33,673 |
| **bn** | 199.83 | 15.19 | 10.03 | 1.19 | 614,374 | 65,553 |
| **gu** | 192.26 | 12.30 | 10.16 | 1.06 | 194,483 | 99,235 |
| **hi** | 354.07 | 18.45 | 11.81 | 1.01 | 1,279,369 | 113,378 |
| **kn** | 150.33 | 11.18 | 8.07 | 1.05 | 399,927 | 39,804 |
| **ml** | 133.79 | 13.96 | 8.82 | 1.10 | 324,994 | 33,314 |
| **mr** | 165.38 | 13.60 | 7.82 | 1.05 | 463,949 | 61,137 |
| **or** | 148.23 | 12.33 | 7.51 | 1.07 | 480,236 | 43,977 |
| **pa** | 215.42 | 8.86 | 12.60 | 1.03 | 343,036 | 47,931 |
| **ta** | 143.83 | 13.41 | 9.21 | 1.44 | 633,888 | 70,919 |
| **te** | 150.60 | 13.79 | 6.74 | 1.10 | 467,958 | 44,622 |
| **average** | 185.40 | 12.86 | 9.16 | 1.10 | 540,781 | 59,413 |

Table 14: Quantitative statistics of the headline generation dataset, focusing on document-title lengths and vocabulary sizes.

| Language | % of novel n-gram | | | | LEAD | EXT-ORACLE |
|---|---|---|---|---|---|---|
| | n=1 | n=2 | n=3 | n=4 | R-L | R-L |
| **XL-Sum En** | 32.22 | 80.99 | 94.57 | 98.06 | 8.24 | 18.82 |
| **as** | 36.86 | 71.53 | 86.98 | 93.97 | 13.95 | 28.56 |
| **bn** | 46.38 | 78.92 | 90.39 | 94.77 | 13.05 | 27.83 |
| **gu** | 40.49 | 73.17 | 86.18 | 91.69 | 16.50 | 29.13 |
| **hi** | 25.93 | 65.62 | 83.67 | 91.54 | 21.67 | 30.51 |
| **kn** | 39.42 | 72.76 | 87.29 | 93.59 | 17.89 | 29.42 |
| **ml** | 40.74 | 65.91 | 75.69 | 78.79 | 30.07 | 39.97 |
| **mr** | 38.39 | 73.49 | 88.33 | 93.89 | 11.32 | 31.94 |
| **or** | 45.60 | 79.61 | 90.23 | 93.94 | 12.79 | 27.73 |
| **pa** | 24.67 | 57.82 | 73.10 | 81.39 | 28.01 | 35.41 |
| **ta** | 42.99 | 70.84 | 82.25 | 87.43 | 31.37 | 37.98 |
| **te** | 44.24 | 78.79 | 91.77 | 96.12 | 17.93 | 30.42 |
| **average** | 38.70 | 71.68 | 85.08 | 90.65 | 19.50 | 31.72 |

Table 15: Qualitative statistics for headline generation dataset, focusing on n-gram overlaps between document and title. Standard baseline scores such as LEAD and EXT-ORACLE are also included.

| Language | Rouge-L | | | | | | | |
| --- | --- | --- | --- | --- | --- | --- | --- | --- |
| | **Monolingual** | | | | **Multilingual** | | | |
| | **No PT** | **mT5** | **SSIB** | **IB** | **No PT** | **mT5** | **SSIB** | **IB** |
| **as** | 66.78 | 30.80 | 68.26 | **71.56** | 52.17 | 33.58 | 46.82 | 44.64 |
| **bn** | 31.30 | 31.54 | 37.95 | **39.17** | 31.15 | 34.21 | 33.49 | 32.60 |
| **gu** | 24.02 | 31.04 | 31.80 | **33.03** | 27.68 | 31.01 | 30.87 | 31.79 |
| **hi** | 25.15 | 32.55 | 34.49 | **34.57** | 29.68 | 32.68 | 33.60 | 32.70 |
| **kn** | 67.34 | 66.67 | 73.19 | 72.35 | 68.71 | **74.49** | 64.50 | 64.10 |
| **ml** | 49.96 | 39.59 | 60.51 | **60.63** | 56.60 | 47.30 | 57.60 | 57.94 |
| **mr** | 30.23 | 32.88 | 40.78 | **41.58** | 30.59 | 36.76 | 33.04 | 34.08 |
| **or** | 13.46 | 21.22 | **23.93** | 21.95 | 17.45 | 21.94 | 23.74 | 21.62 |
| **pa** | 35.29 | 40.13 | 43.14 | **43.81** | 39.11 | 38.91 | 42.12 | 43.25 |
| **ta** | 41.22 | 46.42 | 46.52 | **46.87** | 42.32 | 43.29 | 45.72 | 45.94 |
| **te** | 30.74 | 31.56 | 41.97 | **42.89** | 35.65 | 32.36 | 35.58 | 36.66 |
| average | 37.77 | 36.77 | 45.69 | **46.22** | 39.19 | 38.78 | 40.64 | 40.48 |

Table 16: The table shows Rouge-L scores for the headline generation test sets. We compare between models without pre-training (no PT), IndicBART (IB), separate script IndicBART (SSIB) and mT5 models in monolingual and multilingual settings.

| Input | वाराणसी में खाद्य सुरक्षा एवं औषधि प्रशासन की लापरवाही से व्यापारियों में नाराजगी देखी जा रही है। खाद्य सुरक्षा एवं औषधि प्रशासन का पोर्टल बंद होने से व्यापारियों का पंजीकरण और उन्हें लाइसेंस नहीं मिल पा रहा है। विभाग के कर्मचारियों के अनुसार दवा व्यवसायियों व खाद्य पदार्थ बेचने वाले व्यापारियों को विभाग में पंजीकरण और लाइसेंस लेना अनिवार्य है। इसमें वे दवा व्यवसायी भी शामिल हैं जो दवा के साथ-साथ खाद्य पदार्थ भी बेचते हैं।.... **Transliteration**: vaaraanasee mein khaady suraksha evam aushadhi prashaasan kee laaparavaahee se vyaapaariyon mein naaraajagee dekhee ja rahee hai. khaady suraksha evam aushadhi prashaasan ka portal band hone se vyaapaariyon ka panjeekaran aur unhen license nahin mil pa raha hai. vibhaag ke karmachaariyon ke anusaar dava vyavasaayiyon va khaady padaarth bechane vaale vyaapaariyon ko vibhaag mein panjeekaran aur license lena anivaary hai. isamen ve dava vyavasaayee bhee shaamil hain jo dava ke saath-saath khaady padaarth bhee bechate hain.... **Translation**: In Varanasi, there is displeasure among the traders due to the negligence of the Food Safety and Drug Administration. Due to the closure of the Food Safety and Drug Administration's portal, traders are unable to register and get their license. According to the employees of the department, it is mandatory for drug dealers and traders selling food items to take registration and license in the department. It also includes drug dealers who sell medicine as well as food items.... |
|---|---|
| **Target Output** | खाद्य व दवा व्यापारियों में नाराजगी <br> **Transliteration**: khaady va dava vyaapaariyon mein naaraajagee <br> **Translation:**Discontent among food and drug traders |

| **Monolingual** | **IB** | पोर्टल बंद होने से व्यापारियों को नहीं मिल पा रहे लाइसेंस <br> **Transliteration**: portal band hone se vyaapaariyon ko nahin mil pa rahe license <br> **Translation:**Traders are unable to get license due to closure of portal |
|---|---|---|
| | **No PT** | खाद्य सुरक्षा पोर्टल बंद होने से नहीं मिल रहा लाइसेंस <br> **Transliteration:**khaady suraksha portal band hone se nahin mil raha license <br> **Translation:**License not getting due to closure of food security portal |
| | **SSIB** | वाराणसी में पोर्टल बंद होने से व्यापारियों में नाराजगी <br> **Transliteration**: vaaraanasee mein portal band hone se vyaapaariyon mein naaraajagee <br> **Translation:**Discontent among traders due to closure of portal in Varanasi |
| | **mT5** | वाराणसी पोर्टल बंद होने से व्यापारियों में नाराजगी <br> **Transliteration**: vaaraanasee portal band hone se vyaapaariyon mein naaraajagee <br> **Translation**: Discontent among traders due to closure of Varanasi portal |

| **Multilingual** | **IB** | खाद्य सुरक्षा एवं औषधि प्रशासन की लापरवाही से व्यापारियों में नाराजगी <br> **Transliteration**: khaady suraksha evam aushadhi prashaasan kee laaparavaahee se vyaapaariyon mein naaraajagee <br> **Translation**: Displeasure among traders due to the negligence of Food Safety and Drug Administration |
|---|---|---|
| | **No PT** | वाराणसी में खाद्य पदार्थ बेचने वालों को नहीं मिल पा रहा लाइसेंस <br> **Transliteration**: vaaraanasee mein khaady padaarth bechane vaalon ko nahin mil pa raha license <br> **Translation:**Those who sell food items in Varanasi are not getting license |
| | **SSIB** | खाद्य सुरक्षा एवं औषधि प्रशासन की लापरवाही से कारोबारियों में नाराजगी <br> **Transliteration**: khaady suraksha evam aushadhi prashaasan kee laaparavaahee se kaarobaariyon mein naaraajagee <br> **Translation:**Displeasure among businessmen due to the negligence of Food Safety and Drug Administration |
| | **mT5** | वाराणसी में खाद्य सुरक्षा एवं औषधि प्रशासन की लापरवाही से व्यापारियों में नाराजगी <br> **Transliteration:**vaaraanasee mein khaady suraksha evam aushadhi prashaasan kee laaparavaahee se vyaapaariyon mein naaraajagee <br> **Translation**: Discontent among traders due to negligence of Food Safety and Drug Administration in Varanasi |

Table 17: Model generated output for News Headline Generation

### B.3 Sentence Summarisation

**Dataset Creation and Cleaning:** Since it is a sentence summarization dataset we simply used headline dataset by using first sentence of the article as input and title as its summary. But not all the examples were correlated. Hence, we compute the *overlapping ratio* between the sentence and summary pair. It is similar to headline dataset overlapping ratio, as mentioned in section B.2. This help us in filter out the least overlapping sample from the dataset. Table 18 contains the splits of the sentence summarization dataset.

**Dataset Statistics:** Table 18 shows the count of examples in train/test/validation split for sentence summarization for different languages. The count of examples in the training set ranges from 2.8k for Malayalam to 78.6k for Hindi, with a total count of 326k for all languages. The Gigaword[22] corpus for sentence summarization for English (Rush et al., 2015) corpus is an order of magnitude larger than the total count of examples in our sentence summa-rization dataset. Table 19 shows some quantitative statistics for the sentence summarization dataset. The count of words in title and sentence is comparable to that of English Gigaword corpus. The size of vocabulary of sentence and summary is smaller than that of English Gigaword corpus.

**Results:** We report the *Rouge-L* scores on the decoded test sets for all the models and languages in Table 21.

**Example Outputs:** Tables 22 contains the outputs generated by all the models that we have trained. In the first example, except for monolingual No PT, the output generated by all the models is comparable. In comparing the reference with the output, we see that the meaning is mainly conveyed. The output generated by the No PT and mT5 models is relatively longer than the reference.

---

| Language | Train | Test | Validation |
|---|---|---|---|
| **Gigaword En** | 3,803,957 | 1,951 | 189,651 |
| **as** | 10,812 | 5,452 | 5,232 |
| **bn** | 17,035 | 2,384 | 2,355 |
| **gu** | 54,788 | 8,460 | 8,720 |
| **hi** | 78,669 | 16,778 | 16,893 |
| **kn** | 61,220 | 1,485 | 9,024 |
| **ml** | 2,855 | 1,580 | 1,520 |
| **mr** | 27,396 | 3,348 | 3,282 |
| **or** | 12,065 | 1,440 | 1,539 |
| **pa** | 31,630 | 3,967 | 4,004 |
| **ta** | 23,098 | 2,948 | 2,874 |
| **te** | 7,119 | 862 | 878 |
| **total** | 326,687 | 48,704 | 56,321 |

Table 18: Size of train, test and validation sets in terms of number of samples for sentence summarization. The total size of the dataset, excluding English, is also included.

| Languages | Average words | | Vocabulary Size | |
|---|---|---|---|---|
| | **Sentence** | **Title** | **Sentence** | **Title** |
| **Gigaword En** | 31.35 | 8.23 | 119,500 | 68,882 |
| **as** | 38.43 | 7.61 | 39,915 | 15,965 |
| **bn** | 27.88 | 9.27 | 37,433 | 18,732 |
| **gu** | 31.69 | 10.05 | 75,964 | 40,689 |
| **hi** | 30.70 | 10.04 | 92,168 | 57,724 |
| **kn** | 27.56 | 8.11 | 41,790 | 20,439 |
| **ml** | 18.33 | 9.02 | 21,153 | 12,752 |
| **mr** | 22.90 | 6.93 | 38,883 | 22,644 |
| **or** | 21.78 | 7.56 | 28,116 | 15,183 |
| **pa** | 37.04 | 12.88 | 68,398 | 36,154 |
| **ta** | 19.79 | 9.10 | 47,424 | 32,805 |
| **te** | 19.55 | 6.02 | 36,756 | 18,793 |
| **average** | 26.88 | 8.78 | 48,000 | 26,534 |

Table 19: Quantitative statistics for sentence summarization, focusing on the lengths of the sentence-title pairs in terms of words, as well as the vocabulary sizes.

| Languages | % of novel n-gram | | | | LEAD |
|---|---|---|---|---|---|
| | **n=1** | **n=2** | **n=3** | **n=4** | **R-L** |
| **Gigaword En** | 21.46 | 40.26 | 46.52 | 48.67 | 23.14 |
| **as** | 35.63 | 66.47 | 83.21 | 91.93 | 17.82 |
| **bn** | 40.48 | 66.66 | 81.36 | 89.15 | 32.33 |
| **gu** | 38.07 | 65.19 | 79.42 | 86.39 | 29.63 |
| **hi** | 31.36 | 66.15 | 82.41 | 90.49 | 25.73 |
| **kn** | 37.27 | 67.31 | 83.28 | 91.31 | 16.40 |
| **ml** | 41.82 | 67.00 | 81.93 | 89.87 | 33.67 |
| **mr** | 38.23 | 71.54 | 87.64 | 94.15 | 16.37 |
| **or** | 39.67 | 69.34 | 84.38 | 91.60 | 18.95 |
| **pa** | 32.54 | 60.19 | 73.29 | 81.01 | 31.61 |
| **ta** | 40.55 | 66.78 | 80.75 | 88.12 | 37.41 |
| **te** | 39.79 | 71.65 | 88.19 | 94.62 | 24.64 |
| **average** | 37.76 | 67.12 | 82.35 | 89.88 | 25.87 |

Table 20: Qualitative statistics for sentence summarization, focusing on n-gram overlaps between the sentence-summary pairs. Baseline scores using the first "k" words of the sentence as a summary are also computed.

| Language | Rouge-L | | | | | | | |
|---|---|---|---|---|---|---|---|---|
| | Monolingual | | | | Multilingual | | | |
| | No PT | mT5 | SSIB | IB | No PT | mT5 | SSIB | IB |
| **as** | 31.90 | 42.58 | 62.65 | 60.13 | 57.72 | **70.26** | 62.57 | 59.29 |
| **bn** | 34.70 | 44.19 | 49.46 | 48.89 | 44.55 | **51.93** | 50.60 | 49.29 |
| **gu** | 35.38 | 45.69 | **45.97** | **45.97** | 40.64 | 45.44 | 45.61 | 45.92 |
| **hi** | 42.05 | 44.86 | 44.88 | **45.57** | 41.64 | 44.03 | 45.15 | 45.34 |
| **kn** | 70.64 | 79.69 | 77.16 | 77.40 | 75.16 | **82.71** | 76.33 | 77.32 |
| **ml** | 30.94 | 52.70 | 64.10 | 62.66 | 59.66 | 64.99 | 63.76 | **66.42** |
| **mr** | 26.94 | 44.63 | 45.22 | 45.04 | 39.00 | 44.33 | 45.52 | **46.50** |
| **or** | 24.99 | 44.51 | 47.65 | 42.50 | 36.82 | 44.01 | **49.23** | 43.65 |
| **pa** | 45.65 | 49.47 | 51.54 | 52.11 | 47.08 | 50.38 | 51.26 | **52.22** |
| **ta** | 48.67 | 56.64 | 56.60 | 56.16 | 50.98 | 55.85 | 56.49 | **56.83** |
| **te** | 28.35 | 47.48 | 52.33 | 52.62 | 44.56 | **53.65** | 52.66 | 53.44 |
| average | 38.20 | 50.22 | 54.32 | 53.55 | 48.89 | **55.23** | 54.47 | 54.20 |

Table 21: The table shows Rouge-L scores for the sentence summarization test sets. We compare between models without pre-training (no PT), Indic BART (IB), separate script Indic BART (SSIB) and mT5 models in monolingual and multilingual settings.

| | | |
|---|---|---|
| **Input** | | वाराणसी में खाद्य सुरक्षा एवं औषधि प्रशासन की लापरवाही से व्यापारियों में नाराजगी देखी जा रही है।<br>**Translation:**In Varanasi, there is resentment among the traders due to the negligence of the Food Safety and Drug Admnistration.<br>**Transliteration**: Varanasi mein khadya suraksha evam aushadhi prashaasan ki laparwahi se vyapariyon mein narazgi dekhi ja rahi hai. |
| **Target Output** | | खाद्य व दवा व्यापारियों में नाराजगी<br>**Translation:**Anger among food and drug traders<br>**Transliteration**: Khadya v dawa vyapariyon mein narazgi |
| **Monolingual** | **IB** | खाद्य सुरक्षा एवं औषधि प्रशासन की लापरवाही से व्यापारियों में नाराजगी<br>**Translation:**Discontent among traders due to the negligence of Food Safety and Drug Administration<br>**Transliteration**: Khadya suraksha evam aushadhi prashasan ki laparwahi se vyapariyon mein narazgi. |
| | **No PT** | वाराणसी में खाद्य सुरक्षा एवं औषधि प्रशासन की सख्ती<br>**Translation:**Food Safety and Drug Administration's strictness in Varanasi<br>**Transliteration**: Varanasi mein khadya suraksha evam aushadhi prashasan ki sakhti. |
| | **SSIB** | खाद्य सुरक्षा एवं औषधि प्रशासन की लापरवाही से व्यापारियों में नाराजगी<br>**Translation:**Discontent among traders due to the negligence of Food Safety and Drug Administration<br>**Transliteration**: Khadya suraksha evam aushadi prashasan ki laparwahi se vyapariyon mein narazgi. |
| | **mT5** | वाराणसी में खाद्य सुरक्षा प्रशासन की लापरवाही से व्यापारियों में नाराजगी<br>**Translation:**Discontent among traders due to negligence of food security administration in Varanasi<br>**Transliteration**: Varanasi mein khadya suraksha prashasan ki laparwahi se vyapariyon mein narazgi . |
| **Multilingual** | **IB** | खाद्य सुरक्षा एवं औषधि प्रशासन की लापरवाही से व्यापारियों में नाराजगी<br>**Translation:**Displeasure among traders due to the negligence of Food Safety and Drug Administration<br>**Transliteration**: Khadya suraksha evam aushadhi prashasan ki laparwahi se vyapariyon se narazgi . |
| | **No PT** | वाराणसी में खाद्य सुरक्षा एवं औषधि प्रशासन की लापरवाही से व्यापारियों में नाराजगी<br>**Translation:**Discontent among traders due to the negligence of Food Safety and Drug Administration in Varanasi.<br>**Transliteration:**Varanasi mein khadya suraksha evam aushadhi prashasan ki laparwahi se vyapariyon mein narazgi . |
| | **SSIB** | खाद्य सुरक्षा एवं औषधि प्रशासन की लापरवाही से व्यापारियों में नाराजगी<br>**Translation:**Discontent among traders due to negligence of Food Safety and Drug Administration<br>**Transliteration**: Khadya suraksha evam aushadhi prashasan ki laparwahi se vyapariyon mein narazgi. |
| | **mT5** | वाराणसी में खाद्य सुरक्षा एवं औषधि प्रशासन की लापरवाही से व्यापारियों में नाराजगी<br>**Translation:**Discontent among traders due to the negligence of Food Safety and Drug Administration in Varansi<br>**Transliteration**: Varansi mein khadya suraksha evam aushadhi prashasan ki laparwahi se vyapariyon mein narazgi. |

Table 22: Model generated output for sentence summarization

## B.4 Paraphrasing

**Initial paraphrase extraction:** We use the pivoting approach to extract the initial set of paraphrases. Prior to pivoting, we normalize and tokenize the English sentences using Sacremoses[23] along with removing white spaces to ensure that the same sentences with differing spaces between words become identical. A single paraphrase example is a tuple consisting of $M$ sentences which are all considered to be paraphrases of each other.

**Dataset cleaning:** After paraphrase extraction, we clean it to remove noise. First, we remove the sentences in an example which are exact duplicates that only differ due to tokenization and spelling. For this, we first normalize and tokenize the sentences using the IndicNLP library (Kunchukuttan, 2020)[24]. Then, we remove the punctuation and white spaces from the sentence. If this results in a single sentence in the example, then the example is discarded.

We then randomly select one paraphrase as the input from each group of paraphrases and calculate n-gram overlap for n=1, 2, 3, and 4 between it and the remaining sentences which are considered as references. We eliminate the paraphrases which have an n-gram overlap ratio greater than 0.8 to ensure that the paraphrases are not very similar. The ratio is calculated using the formula:

$$a_n = \frac{O_n}{I_n}$$
$$b_n = \frac{O_n}{R_n}$$
$$ratio = \frac{\sum_{n=1}^{4} \frac{2}{\frac{1}{a_n} + \frac{1}{b_n}}}{4}$$

where $O_n$=n-gram overlap between reference and input, $I_n$=Total n-grams in input and $R_n$=Total n-grams in reference. This formula computes the average of 1-, 2-, 3-, and 4-gram overlaps. These overlaps are computed as the harmonic mean between the overlapping n-grams relative to the input ($a_n$) and the reference ($b_n$). This ensures that the overlap information is not biased towards either the input or the reference.

Next for each input, we select up to five references. We first sort the references based on the n-gram overlap scores with respect to the input, then divide the scores into 5 equal intervals and finally, select the example corresponding to the first score in each interval. If for a particular input, the number of references is less than 5, we keep all the references.

**Dataset splitting:** We split the collection of examples into train, validation and test sets. We first sort examples in the descending order of the number of paraphrases in them, the top 10,000 go into the test set, the next 10,000 into the validation set and the remaining into the training set. Except Assamese, all languages have 10,000 examples with 5 references[25] per input in the validation and test set. The training set has anywhere between 1 and 5 references. Assamese, due to its low-resource nature, could only be split into validation and test sets with 4,420 examples each.

The per language statistics are given in Table 23. We compare our statistics with the English language portion of OpusParcus corpus (Creutz, 2018).

---

| Language | Train | | Test | | Validation | |
|---|---|---|---|---|---|---|
| | #Instances | #Total Paraphrases | #Instances | #Total Paraphrases | #Instances | #Total Paraphrases |
| **OpusParcus En** | - | - | 1,445 | 2,890 | - | - |
| **as** | - | - | 4,420 | 12,965 | 4,420 | 8,840 |
| **bn** | 890,445 | 2,837,641 | 10,000 | 60,000 | 10,000 | 60,000 |
| **gu** | 379,202 | 1,145,079 | 10,000 | 60,000 | 10,000 | 60,000 |
| **hi** | 929,507 | 2,690,315 | 10,000 | 60,000 | 10,000 | 60,000 |
| **kn** | 522,148 | 1,683,599 | 10,000 | 60,000 | 10,000 | 60,000 |
| **ml** | 761,933 | 2,426,155 | 10,000 | 60,000 | 10,000 | 60,000 |
| **mr** | 406,003 | 1,263,050 | 10,000 | 60,000 | 10,000 | 60,000 |
| **or** | 105,970 | 265,360 | 10,000 | 60,000 | 10,000 | 56,020 |
| **pa** | 266,704 | 734,573 | 10,000 | 60,000 | 10,000 | 60,000 |
| **ta** | 497,798 | 1,631,560 | 10,000 | 60,000 | 10,000 | 60,000 |
| **te** | 596,283 | 1,977,942 | 10,000 | 60,000 | 10,000 | 60,000 |
| **total** | 5,355,993 | 16,655,274 | 104,420 | 612,965 | 104,420 | 604,860 |

Table 23: Language-Wise Statistics for paraphrase generation dataset. Each instance consist of up to 6 paraphrases, of which one is chosen as the input and the rest are references, ordered according to the least n-gram overlap.

**Dataset statistics:** Table 24 gives the percentage of novel n-grams, i.e. n-grams in the reference, not present in the input. The statistics show high percentage of novel n-grams for higher n-grams (n=2,3,4). We compare the statistics with the English portion of the OpusParcus Test set (Creutz, 2018). The statistics are comparable between English and Indic languages.

**Results**: We report the BLEU, Self-BLEU and iBLEU scores on the test set for all the languages in Tables 25- 27.

**Example Outputs**: Table 28 contains the output generated by all the models that we have trained. In the first example in Table 28 the No PT+multilingual model only makes trivial changes to the input. On the other hand, the output generated by the model IB+Multilingual is diverse. The pre-trained models generate better paraphrases as they provide more context as to what level we are talking about the rank of the university; however, the output generated by the No PT model doesn't give information on the level.

| Language | % of novel n-grams | | | |
|---|---|---|---|---|
| | n=1 | n=2 | n=3 | n=4 |
| **OpusParcus En** | 63.22 | 92.23 | 98.24 | 99.72 |
| **bn** | 70.83 | 88.18 | 93.39 | 95.36 |
| **gu** | 62.90 | 86.08 | 93.45 | 96.42 |
| **hi** | 52.99 | 75.58 | 85.44 | 90.49 |
| **kn** | 73.51 | 91.64 | 96.48 | 97.98 |
| **ml** | 75.93 | 92.13 | 96.32 | 97.75 |
| **mr** | 66.19 | 86.68 | 93.52 | 96.44 |
| **or** | 63.58 | 85.35 | 92.96 | 96.20 |
| **pa** | 56.65 | 79.81 | 89.08 | 93.39 |
| **ta** | 73.59 | 90.66 | 95.44 | 97.15 |
| **te** | 72.20 | 90.58 | 96.00 | 97.81 |
| **average** | 66.84 | 86.67 | 93.21 | 95.90 |

Table 24: Percentage of novel n-grams in references not present in the input for paraphrase generation.

| | BLEU ↑ | | | | | | | |
|---|---|---|---|---|---|---|---|---|
| Language | Monolingual | | | | Multilingual | | | |
| | No PT | IB | SSIB | mT5 | No PT | IB | SSIB | mT5 |
| **as** | 1.02 | 1.12 | 0.69 | 0.94 | 1.61 | **1.66** | 1.19 | 1.11 |
| **bn** | 10.60 | 11.30 | 10.15 | 4.44 | 8.85 | **11.57** | 10.04 | 4.45 |
| **gu** | 18.48 | 21.14 | 16.80 | 7.36 | 18.53 | **22.10** | 18.69 | 8.66 |
| **hi** | 24.94 | **27.55** | 25.35 | 12.50 | 22.94 | 27.29 | 25.05 | 13.77 |
| **kn** | 13.56 | 15.25 | 12.83 | 7.11 | 12.93 | **15.40** | 13.14 | 7.52 |
| **ml** | 9.96 | 10.36 | 8.68 | 7.81 | 9.48 | **10.57** | 8.71 | 7.75 |
| **mr** | 17.86 | 17.74 | 18.56 | 7.52 | 17.48 | **20.38** | 18.50 | 8.64 |
| **or** | 11.21 | 15.34 | 20.84 | 0.50 | 16.59 | 19.26 | **23.02** | 1.77 |
| **pa** | 12.93 | 14.44 | **18.01** | 7.39 | 12.40 | 14.87 | 17.61 | 8.72 |
| **ta** | 15.51 | 18.04 | 16.03 | 12.20 | 15.52 | **18.52** | 16.25 | 12.39 |
| **te** | 15.16 | **16.78** | 13.69 | 7.48 | 13.54 | 16.70 | 14.16 | 8.16 |
| **average** | 13.75 | 15.37 | 14.69 | 6.84 | 13.62 | **16.21** | 15.12 | 7.54 |

Table 25: The table shows BLEU scores for the paraphrase generation test sets. We compare between models without pre-training (no PT), Indic BART (IB), separate script Indic BART (SSIB) and mT5 models in monolingual and multilingual settings.

| | Self-BLEU ↓ | | | | | | | |
|---|---|---|---|---|---|---|---|---|
| Language | Monolingual | | | | Multilingual | | | |
| | No PT | mT5 | SSIB | IB | No PT | mT5 | SSIB | IB |
| **as** | 1.22 | 1.07 | **0.83** | 1.3 | 2.29 | 1.18 | 1.64 | 2.06 |
| **bn** | 1.2 | 0.19 | **0.88** | 1.32 | 2.18 | 0.1 | 1.08 | 1.69 |
| **gu** | 1.84 | **0.49** | 1.96 | 2.39 | 3.14 | **0.49** | 1.62 | 2.76 |
| **hi** | 2.66 | 0.56 | 1.52 | 2.44 | 2.89 | **0.52** | 1.75 | 2.87 |
| **kn** | 2.47 | 1.18 | 1.72 | 2.75 | 3.66 | **0.87** | 1.89 | 2.98 |
| **ml** | 1.55 | 0.64 | 1.01 | 1.33 | 2.03 | **0.58** | 1.36 | 1.7 |
| **mr** | 1.28 | **0.28** | 0.83 | 1.18 | 2.77 | 0.31 | 1.49 | 2.2 |
| **or** | 0.99 | **0.01** | 1.28 | 1.36 | 2.67 | **0.01** | 2.68 | 2.1 |
| **pa** | 0.97 | **0.32** | 1.46 | 1.2 | 1.64 | 0.35 | 1.37 | 1.35 |
| **ta** | 2.56 | 1.49 | 1.96 | 2.29 | 3.45 | **1.22** | 2.13 | 2.88 |
| **te** | 2.25 | 0.62 | 1.31 | 2.28 | 3.72 | **0.71** | 2.29 | 3.34 |
| **average** | 1.73 | 0.62 | 1.34 | 1.80 | 2.77 | **0.58** | 1.75 | 2.36 |

Table 26: The table shows Self-BLEU scores for the paraphrase generation test sets. We compare between models without pre-training (no PT), IndicBART (IB), separate script IndicBART (SSIB) and mT5 models in monolingual and multilingual settings.

| Language | iBLEU ↑ | | | | | | | |
|---|---|---|---|---|---|---|---|---|
| | Monolingual | | | | Multilingual | | | |
| | No PT | mT5 | SSIB | IB | No PT | mT5 | SSIB | IB |
| as | 0.35 | 0.34 | 0.23 | 0.39 | 0.44 | 0.42 | 0.34 | **0.54** |
| bn | 7.06 | 3.05 | 6.84 | 7.51 | 5.54 | 3.09 | 6.7 | **7.59** |
| gu | 12.38 | 5.01 | 11.17 | 14.08 | 12.03 | 5.92 | 12.6 | **14.64** |
| hi | 16.66 | 8.58 | 17.29 | **18.55** | 15.19 | 9.48 | 17.01 | 18.24 |
| kn | 8.75 | 4.62 | 8.47 | 9.85 | 7.95 | 5 | 8.63 | **9.89** |
| ml | 6.51 | 5.28 | 5.77 | 6.85 | 6.03 | 5.25 | 5.69 | **6.89** |
| mr | 12.12 | 5.18 | 12.74 | 12.06 | 11.41 | 5.96 | 12.5 | **13.61** |
| or | 7.55 | 0.35 | 14.2 | 10.33 | 10.81 | 1.24 | **15.31** | 12.85 |
| pa | 8.76 | 5.08 | **12.17** | 9.75 | 8.19 | 6 | 11.92 | 10 |
| ta | 10.09 | 8.09 | 10.63 | 11.94 | 9.83 | 8.31 | 10.74 | **12.10** |
| te | 9.94 | 5.05 | 9.19 | **11.06** | 8.36 | 5.5 | 9.23 | 10.69 |
| average | 9.11 | 4.60 | 9.88 | 10.22 | 8.71 | 5.11 | 10.06 | **10.64** |

Table 27: The table shows iBLEU scores for the paraphrase generation test sets. We compare between models without pre-training (no PT), Indic BART (IB), separate script Indic BART (SSIB) and mT5 models in monolingual and multilingual settings.

| | | |
|---|---|---|
| **Input** | दिल्ली यूनिवर्सिटी देश की प्रसिद्ध यूनिवर्सिटी में से एक है.<br>**Translation:**Delhi University is one of the famous universities of the country.<br>**Transliteration:** Delhi university desh ki prasidha university mein se ek hai. | |
| **Target Output** | दिल्ली विश्वविद्यालय, भारत में उच्च शिक्षा के लिए एक प्रतिष्ठित संस्थान है।<br>**Translation:**The University of Delhi is a prestigious institution for higher education in India.<br>**Transliteration:** Delhi Vishwavidyalaya, Bharat mein uchch shiksha ke liye ek pratishthit sansthaan hai. | |
| **Monolingual** | **IB** | दिल्ली विश्वविद्यालय देश का सबसे बड़ा विश्वविद्यालय है।.<br>**Translation:**Delhi University is the largest university in the country.<br>**Transliteration:** Delhi vishwavidyalaya desh ka sabse bada vishwavidyalaya hai. |
| | **No PT** | दिल्ली यूनिवर्सिटी एक प्रसिद्ध विश्वविद्यालय है।<br>**Translation:**Delhi University is a famous university.<br>**Transliteration:** Delhi university ek prasidha vishwavidyalaya hai. |
| | **SSIB** | दिल्ली विश्वविद्यालय देश की सबसे बड़ी विश्वविद्यालयों में शामिल है।<br>**Translation:**Delhi University is one of the largest universities in the country.<br>**Transliteration:** Delhi vishwavidyalaya desh ki sabse badi vishwavidyalayon mein shamil hai. |
| | **mT5** | दिल्ली विश्वविद्यालय देश की सबसे बड़ी विश्वविद्यालय है।<br>**Translation:**Delhi University is the largest university in the country.<br>**Transliteration:** Delhi vishwavidyalaya desh ki sabse badi vishwavidyalaya hai. |
| **Multilingual** | **IB** | दिल्ली विश्वविद्यालय देश की प्रमुख विश्वविद्यालयों में शामिल है।<br>**Translation:**Delhi University is one of the leading universities in the country.<br>**Transliteration:** Delhi vishwavidyalaya desh ki pramukh vishwavidyalayon mein shamil hai. |
| | **No PT** | दिल्ली यूनिवर्सिटी भारत के प्रसिद्ध यूनिवर्सिटी से एक हैं।<br>**Translation:**Delhi University is one of the famous universities of India.<br>**Transliteration:** Delhi university bharat ke prasidha university mein se ek hai. |
| | **SSIB** | दिल्ली यूनिवर्सिटी भारत की सबसे बड़ी यूनिवर्सिटी है।<br>**Translation:**Delhi University is the largest university in India.<br>**Transliteration:** Delhi university bharat ki sabse badi university mein hai. |
| | **mT5** | दिल्ली विश्वविद्यालय देश की सबसे बड़ी विश्वविद्यालय है।<br>**Translation:**Delhi University is the largest university in the country.<br>**Transliteration:** Delhi vishwavidyalaya desh ki sabse badi vishwavidyalaya hai. |

Table 28: Model generated output for paraphrase generation

## B.5  Question Generation

**Dataset extraction From SQuAD dataset:** Table 29 shows the original sample, in which 'Context' is a paragraph which is associated with multiple question and answer pairs. Each 'Answer' has an 'Answer Text' which is the actual answer and a 'Answer Start Index' which is the index of the first character of 'Answer Text' in 'context'. The bold text in the Context row of the Original English Sample is the sentence which contains the answer to the question. This sentence is extracted with the help of 'Answer Start Index' for each question-answer pair and is treated as a reference context.

Finally, each sample of the English dataset has <Context, Answer, Question, ID> and then this is translated using IndicTrans[26] to the required Indic language dataset.The example of the translated dataset in Hindi and Marathi is shown in the same table. We use SQuAD-v1 train[27] set for training and validation set, where the split is 80-20 respectively. We use SQuAD-v1 dev[28] set as test set directly. The train, dev and test splits contain 69,979, 17,495, and 10,553 examples, respectively.

**Dataset statistics:** Table 30 gives the novel n-gram percentage between question and context. We can see that Indic languages have higher novel n-grams when compared to the English dataset.

**Results**: We report the Rouge-L scores on the test set for all the 11 languages in Table 31.

**Output:** Table 32 shows the output generated by all the eight models. We show the example in the native language (Hindi here), but we also mention translation and transliteration of the native text in English for better understanding. The target output is about who won the match and, output generated by monolingual IB, multilingual mT5, multilingual IB and, multilingual No PT are very close to the target output. In contrast, other outputs is more about who lost the match. But all of these outputs are related to the actual context and the answer.

---

| Original English sample | |
|---|---|
| Context | Architecturally, the school has a Catholic character. Atop the Main Building's gold dome is a golden statue of the Virgin Mary. Immediately in front of the Main Building and facing it, is a copper statue of Christ with arms upraised with the legend Venite Ad Me Omnes. Next to the Main Building is the Basilica of the Sacred Heart. Immediately behind the basilica is the Grotto, a Marian place of prayer and reflection. **It is a replica of the grotto at Lourdes, France, where the Virgin Mary reputedly appeared to Saint Bernadette Soubirous in 1858.** At the end of the main drive (and in a direct line that connects through 3 statues and the Gold Dome), is a simple, modern stone statue of Mary. |
| Question | To whom did the Virgin Mary allegedly appear in 1858 in Lourdes, France? |
| Answer Text | Saint Bernadette Soubirous |
| Answer Start index | 515 |
| ID | 5733be284776f41900661182 |
| **Converted English sample** | |
| Context | It is a replica of the grotto at Lourdes, France, where the Virgin Mary reputedly appeared to Saint Bernadette Soubirous in 1858. |
| Answer | Saint Bernadette Soubirous |
| Question | To whom did the Virgin Mary allegedly appear in 1858 in Lourdes, France? |
| ID | 5733be284776f41900661182 |
| **Hindi Translated sample** | |
| Context | यह लूर्देस, फ्रांस में स्थित ग्रॉटो की प्रतिकृति है, जहां 1858 में सेंट बनडिट सौबिरस को वर्जिन मैरी दिखाई दी थी। (R:yah loordes, phraans mein sthit groto kee pratikrti hai, jahaan 1858 mein sent barnaadet saubiras ko varjin mairee dikhaee dee thee.) |
| Answer | संत बनडिट साबिरोस (R:sant barnaadet saabiros) |
| Question | सन् 1858 में लूईस फ्रांस में कुँवारी मरियम कथित तौर पर किसके सामने प्रकट हुई? (R:san 1858 mein loordas phraans mein kunvaaree mariyam kathit taur par kisake saamane prakat huee?) |
| ID | 5733be284776f41900661182 |

Table 29: Translation example for question generation. Here R stands for Romanisation, that is transliteration in English of the native text.

| Language | Question And Context Overlap | | | |
|---|---|---|---|---|
| | n=1 | n=2 | n=3 | n=4 |
| **en** | 63.33 | 85.35 | 92.40 | 95.61 |
| **as** | 78.21 | 93.01 | 97.51 | 98.96 |
| **bn** | 74.22 | 91.20 | 96.75 | 98.66 |
| **gu** | 76.95 | 92.18 | 97.04 | 98.78 |
| **hi** | 62.61 | 85.17 | 93.15 | 96.53 |
| **kn** | 78.10 | 92.91 | 97.46 | 98.95 |
| **ml** | 79.80 | 93.33 | 97.68 | 99.11 |
| **mr** | 78.66 | 93.04 | 97.45 | 98.98 |
| **or** | 76.62 | 92.53 | 97.24 | 98.87 |
| **pa** | 66.76 | 87.57 | 94.63 | 97.47 |
| **ta** | 80.15 | 93.69 | 97.83 | 99.16 |
| **te** | 76.58 | 92.22 | 97.21 | 98.90 |
| **average** | 75.33 | 91.53 | 96.72 | 98.58 |

Table 30: The table shows the percentage of novel n-gram for question generation. We give the statistics of novel n-gram in question string compared to context string.

| Language | Rouge-L | | | | | | | |
|---|---|---|---|---|---|---|---|---|
| | Monolingual | | | | Multilingual | | | |
| | No PT | mT5 | SSIB | IB | No PT | mT5 | SSIB | IB |
| as | 11.89 | 19.69 | 20.33 | 20.21 | 15.36 | **21.01** | 20.73 | 20.48 |
| bn | 14.46 | 29.56 | 26.61 | 24.49 | 20.63 | **30.58** | 30.38 | 26.63 |
| gu | 15.92 | 26.31 | 25.24 | 26.25 | 22.53 | 27.29 | **28.13** | 27.71 |
| hi | 21.67 | 34.58 | 33.60 | 32.24 | 29.14 | **35.49** | 34.42 | 35.38 |
| kn | 11.31 | 23.32 | 21.32 | 22.40 | 18.18 | **24.24** | 23.77 | 23.56 |
| ml | 9.94 | 21.82 | 19.87 | 19.71 | 16.93 | **22.30** | 22.24 | 22.17 |
| mr | 12.18 | 22.81 | 21.13 | 20.61 | 18.37 | **23.93** | 23.62 | 23.52 |
| or | 14.87 | 20.34 | 25.70 | 24.29 | 20.80 | 22.11 | **27.53** | 25.25 |
| pa | 18.91 | 29.72 | 30.74 | 30.59 | 26.08 | 30.97 | **32.53** | 32.10 |
| ta | 5.09 | 22.84 | 21.24 | 21.24 | 12.01 | **23.63** | 23.49 | 22.98 |
| te | 12.38 | 25.63 | 23.15 | 24.46 | 19.71 | **26.37** | 25.81 | 25.67 |
| average | 13.51 | 25.15 | 24.45 | 24.23 | 19.98 | 26.17 | **26.60** | 25.95 |

Table 31: Rouge-L scores for question generation on the synthetic dataset. We compare between models without pretraining (no PT), IndicBART (IB), separate script IndicBART (SSIB) and mT5 models in monolingual and multilingual settings.

| Input | डेनवर ब्रॉंकोस <DELIM> अमेरिकी फुटबॉल सम्मेलन (एएफसी) के चैंपियन डेनवर ब्रॉंकोस ने नेशनल फुटबॉल कांफ्रेंस (एनएफसी) की चैंपियन कैरोलिना पैंथर्स को 24-10 से हराकर अपना तीसरा सुपर बाउल खिताब जीता। <br> **Translation**: Denver Broncos <DELIM> American Football Conference (AFC) champion Denver Broncos defeated National Football Conference (NFC) champion Carolina Panthers 24-10 to win their third Super Bowl title. <br> **Transliteration**: denavar bronkos <dailim> ameriket phutabol sammelan (eephasee) ke chaimpiyan denavar bronkos ne neshanal phutabol kaamphrens (enephasee) kee chaimpiyan kairolina painthars ko 24-10 se haraakar apana teesara supar baul khitaab jeeta. | | |
|---|---|---|---|
| Target Output | कौन सी एनएफएल टीम ने सुपर बाउल 50 जीता? <br> **Translation**:Which NFL team won Super Bowl 50? <br> **Transliteration**: kaun see enephel teem ne supar baul 50 jeeta? | | |
| Monolingual | No PT | कौन सा अमेरिकी फुटबॉल सम्मेलन फीफा विश्व कप हार गया? <br> **Translation**: Which American Football Conference lost the FIFA World Cup? <br> **Transliteration**: kaun sa ameriket phutabol sammelan pheepha vishv kap haar gaya? |
| | IB | नॉर्थ कैरोलिना पैंथर्स को हराकर अपना तीसरा सुपर बाउल खिताब किसने जीता? <br> **Translation**: Who defeated the North Carolina Panthers to win their third Super Bowl title? <br> **Transliteration**: north kairolina painthars ko haraakar apana teesara supar baul khitaab kisane jeeta? |
| | SSIB | एएफसी का चैंपियन कौन है? कौन है? <br> **Translation**:Who is the champion of AFC? Who is it? <br> **Transliteration**: eephasee ka chaimpiyan kaun hai? kaun hai? |
| | mT5 | नेशनल फुटबॉल कांफ्रेंस की चैंपियन कैरोलिना पैंथर्स को किसने हराया? <br> **Translation**: Who defeated the National Football Conference champion Carolina Panthers? <br> **Transliteration**: neshanal phutabol kaamphrens kee chaimpiyan kairolina painthars ko kisane haraaya? |
| Multilingual | No PT | कौन सी टीम ने नेशनल फुटबॉल सम्मेलन को अपना तीसरा खिताब जीता? <br> **Translation**: Which team won the National Football Conference its third title? <br> **Transliteration**: kaun see teem ne neshanal phutabol sammelan ko apana teesara khitaab jeeta? |
| | IB | नॉर्थ कैरोलिना पैंथर्स को हराकर तीसरा सुपर बाउल खिताब किसने जीता? <br> **Translation**:Who defeated the North Carolina Panthers to win the third Super Bowl title? <br> **Transliteration**: north kairolina painthars ko haraakar teesara supar baul khitaab kisane jeeta? |
| | SSIB | कौन सा एएफसी चैंपियन कैरोलिना पैंथर्स को 24-10 से हराया? <br> **Translation**: Which AFC champions Carolina Panthers beat 24-10? <br> **Transliteration**: kaun sa eephasee chaimpiyan kairolina painthars ko 24-10 se haraaya? |
| | mT5 | अमेरिकी फुटबॉल सम्मेलन के किस चैंपियन ने तीसरा सुपर बाउल खिताब जीता? <br> **Translation**: Which champion of the American Football Conference won the third Super Bowl title? <br> **Transliteration**: ameriket phutabol sammelan ke kis chaimpiyan ne teesara supar baul khitaab jeeta? |

Table 32: Model generated output for Question Generation. We compare between models without pretraining (no PT), IndicBART (IB), separate script IndicBART (SSIB) and mT5 models in monolingual and multilingual settings.